\documentclass[conference]{IEEEtran}
\usepackage{times}

\usepackage[numbers]{natbib}
\usepackage{multicol}
\usepackage[bookmarks=true]{hyperref}

\usepackage{amsmath,amssymb}
\usepackage{graphicx}
\usepackage{enumerate}
\usepackage[dvipsnames]{xcolor}
\usepackage{multirow}
\usepackage[caption=false, font=footnotesize]{subfig}
\usepackage{caption}
\usepackage{bm}
\usepackage{algorithm}
\usepackage{algorithmic}
\usepackage{balance}

\newcommand{\R}{{\mathbb{R}}}

\newcommand{\D}{\mathrm{d}}
\newcommand{\Log}{\mathrm{Log}}
\newcommand{\Exp}{\mathrm{Exp}}
\newcommand{\ad}{\mathrm{ad}}
\newcommand{\fgfunc}{f_g(b_t^g, \tilde{\omega}_t,\dot{\tilde{\omega}}_t)}
\newcommand{\fafunc}{f_a(b_t^a, \tilde{a}_t,\dot{\tilde{a}}_t)}
\newcommand{\fgfunctheta}{f_g(b_t^g, \tilde{\omega}_t,\dot{\tilde{\omega}}_t;\theta)}
\newcommand{\fafunctheta}{f_a(b_t^a, \tilde{a}_t,\dot{\tilde{a}}_t;\theta)}

\newtheorem{prob}{\textbf{Problem}}
\newtheorem{rmk}{\textbf{Remark}}
\newtheorem{thm}{\textbf{Theorem}}

\begin{document}

\title{Debiasing 6-DOF IMU via Hierarchical Learning of Continuous~Bias~Dynamics}




%
\author{\authorblockN{Ben Liu\textsuperscript{1}, 
Tzu-Yuan Lin\textsuperscript{2}, 
Wei Zhang\textsuperscript{1}\authorrefmark{2},  and 
Maani Ghaffari\textsuperscript{2}}
\authorblockA{\authorrefmark{2}Corresponding Author}
\authorblockA{\textsuperscript{1}Southern University of Science and Technology
\\ Email: liub2021@mail.sustech.edu.cn; zhangw3@sustech.edu.cn}
\authorblockA{\textsuperscript{2}University of Michigan\\
Email: tzuyuan@umich.edu; maanigj@umich.edu}
}

\maketitle

\begin{abstract}
This paper develops a deep learning approach to the online debiasing of IMU gyroscopes and accelerometers. Most existing methods rely on implicitly learning a bias term to compensate for raw IMU data. Explicit bias learning has recently shown its potential as a more interpretable and motion-independent alternative. However, it remains underexplored and faces challenges, particularly the need for ground truth bias data, which is rarely available. To address this, we propose a neural ordinary differential equation (NODE) framework that explicitly models continuous bias dynamics, requiring only pose ground truth, often available in datasets. This is achieved by extending the canonical NODE framework to the matrix Lie group for IMU kinematics with a hierarchical training strategy. The validation on two public datasets and one real-world experiment demonstrates significant accuracy improvements in IMU measurements, reducing errors in both pure IMU integration and visual-inertial odometry. 
\end{abstract}

\IEEEpeerreviewmaketitle




\section{Introduction}
Inertial Measurement Units (IMUs) are essential in robotic applications, providing angular velocity and acceleration measurements that support various state estimation tasks. A notable use of IMUs is in Visual-Inertial Odometry (VIO) \cite{mourikis2007MSCKF,qin2018vinsmono,qin2019Vinsfusion_general,geneva2020openvins,vanGoor2023EqVIO}, where IMU and camera data are fused to estimate a robot’s orientation, velocity, and position. However, low-cost IMUs are often prone to significant noise and bias~\cite{novatel_apn064}, leading to inaccurate measurements that can compromise VIO performance. This becomes even more critical in scenarios where cameras fail due to adverse environmental conditions \cite{buchanan2022deepbias,lin2021leggedcontactlearning}, leaving the IMU as the sole source of odometry information. In such scenarios, the odometry output heavily depends on the accuracy of IMU data. 
Therefore, deriving accurate measurements from raw IMU data that is noisy and biased is critical for robust state estimation.

To address the inaccuracies inherent in raw IMU data, \emph{calibration} methods are utilized to enhance measurement accuracy for a specific IMU device \cite{chen2024deepimusurvey}. The typical method uses a linear model to model axis misalignment and scale factors, while the bias is often modeled as a constant or a Brownian motion process \cite{rehder2016kalibr,schubert2018tum}. However, the IMU model is complex and difficult to represent accurately, particularly because the bias can be time-varying and influenced by factors such as temperature and vibration. With the advancement of deep learning, researchers have begun leveraging neural networks to calibrate IMUs, aiming to capture ignored or simplified components in model-based methods. \citet{brossard2020denoising} employs a convolutional neural network to directly learn a correction term for the gyroscope based on local windows of IMU data. 
Similar methods have been extended to both gyroscopes and accelerometers \cite{liu2023DUET}. However, it has been noted that such methods may not effectively distinguish whether the network learns deviation characteristics from the motion pattern or from the IMU itself \cite{buchanan2022deepbias}. To address this issue, \citet{buchanan2022deepbias} proposed explicitly modeling the IMU bias evolution using a neural network to achieve motion-independent calibration. However, this method requires ground truth bias data from sensor fusion with other sensors like LiDAR or cameras. The accuracy of bias estimation thus depends heavily on the fusion algorithm, and the efforts to obtain the bias ground truth are non-trivial. 
\begin{figure}[t]
    \centering
    \includegraphics[width=\linewidth]{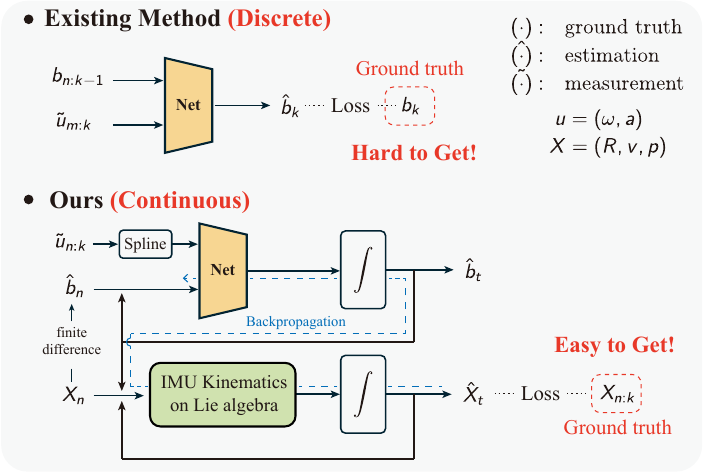}
    \caption{Training process for the explicit evolution of bias. The subscript notation $u_{n:k}$ represents $u_n,u_{n+1},...,u_k$. The existing method \cite{buchanan2022deepbias} models bias evolution using a discrete approach, which requires ground-truth bias values during training. In contrast, our method employs a continuous model to capture bias dynamics and does not rely on ground-truth bias for training. 
    }
    \label{fig:first-fig}
    \vspace{-10px}
\end{figure}

In this work, we propose a framework to model the continuous dynamics of the bias \emph{without} the need for ground truth bias data.
The contributions of this work can be summarized as follows: 
\begin{enumerate}[1.]
    \item We propose a novel loss formulation that bypasses the requirement for bias ground truth during training. Moreover, its hierarchical structure makes network design and tuning more efficient. 
    \item We model the bias dynamics as a vector field in the Lie algebra with a learning approach, allowing us to utilize canonical neural ordinary differential equation tools effectively.
    The vector field formulation allows us to obtain continuous dynamics of the bias, resulting in lightweight networks that deliver superior performance. 
    \item We evaluate our method on two public datasets and one real-world experiment, demonstrating that the proposed method provides more accurate IMU measurements than existing methods and has generalization capability. 
    \item We offer an open-source project available at {\small\url{https://github.com/UMich-CURLY/Debias_IMU.git}}.
\end{enumerate}

\section{Related Work}
We review some learning-based methods in IMU estimation and their relationship with our method.
\subsection{Learning Motion Pattern from IMU}
One direction is to extract the motion pattern information from IMU data. 
Early work \citet{yan2018ridi} utilized a combination of Support Vector Machine (SVM) and Support Vector Regression (SVR) to predict a person's motion displacement based on local windowed IMU data. Building on this approach, the same authors later employed a neural network to perform a similar task \cite{yan2019ronin}. Both studies only focused specifically on 2D motion scenarios. Subsequently, \citet{liu2020tlio} advanced this line of research by integrating the learned displacement into an Extended Kalman Filter (EKF), using the displacement estimates as updates with IMU-based predictions. These methods assume that motion within a given category exhibits finite patterns that can be effectively regressed using IMU data. However, these approaches require large datasets for training and are sensitive to the motion context \cite{buchanan2022deepbias}; for instance, a model trained on flat-ground motion may not generalize well to stair-climbing scenarios. Therefore, when the training data is not sufficient and motion contexts are complex, device-specific calibration is necessary. In addition, some of these methods use calibrated IMU data as input \cite{yan2019ronin,liu2020tlio}, focusing on incorporating the motion information to improve estimation accuracies. On the contrary, our method focuses on denoising and debiasing the raw IMU measurements.


\subsection{Learning Calibration Model for IMU}
Another line of research assumes that neural networks can provide accurate IMU measurements from raw inaccurate data, typically restricted to a single device. These refined IMU measurements can be used in inertial-based odometry, such as VIO, or to learn motion patterns as in prior methods.

Esfahani et al. \cite{esfahani2019orinet} introduced a Long Short-Term Memory (LSTM) framework to denoise gyroscope data and directly estimate orientation, where the orientation integration process is implicitly learned by the neural network. Similarly, Sun et al. \cite{sun2021IDOL} applied an LSTM to predict both orientation and position. To improve short-term accuracy, they incorporated an extra EKF that uses raw IMU data to correct orientation, observing that the LSTM performs well over long durations but poorly in shorter timeframes. These methods combine denoising and integration into a single framework, which is hard to separate. In contrast, Brossard et al. \cite{brossard2020denoising} focused solely on gyroscope denoising by modeling bias and noise as a single term. By compensating for this term, the clean angular velocity can be obtained and can subsequently be numerically integrated into orientation. Their approach uses a dilated convolutional neural network that takes windowed IMU data as input. Building on this framework, Huang et al. \cite{huang2022memsIMUTCN} explored the use of a temporal convolutional network structure and demonstrated improved performance, while Liu et al. \cite{liu2023DUET} extended the approach to handle both gyroscope and accelerometer data simultaneously. To ensure the generality, Zhang et al. \cite{zhang2021imuRNNintgtalterm} construct the loss function based on partial integration terms and employ a recurrent neural network to estimate accurate IMU measurements. All these methods implicitly learn the noise or bias of IMU data, regardless of whether the numerical integration is explicitly processed or implicit embedded in the neural network.

More recently, Buchanan et al. \cite{buchanan2022deepbias} proposed explicitly modeling bias evolution using a transformer or LSTM, which is embedded within a Maximum-a-Posteriori (MAP) framework alongside other sensor inputs. While this explicit bias modeling demonstrates generalizability across motion patterns, it relies on accurate bias ground truth derived from additional sensor fusion algorithms, which are challenging to obtain and sensitive to the fusion method.

In this work, we also focus on explicitly learning the bias evolution process but with a different approach: modeling the continuous dynamics of the bias. This framework reduces neural network complexity and eliminates the need for bias ground truth, making it lightweight and more practical.

\section{Problem Statement}

\subsection{IMU Model}
An IMU measures angular velocity, denoted as $\omega_t$, and linear acceleration (include gravity), 
denoted as $a_t$, both expressed in the IMU frame. A commonly adopted measurement model for these quantities is presented in \cite{bloesch2013EKF,hartley2020InEKF, kim2021SEofleg_RAL}:
\begin{equation}
    \tilde{\omega}_t = \omega_t + b_t^g + n_t^g, \quad \tilde{a}_t = a_t + b_t^a + n_t^a,
    \label{eq:2}
\end{equation}
where the superscript $\tilde{(\cdot)}$ indicates measured quantities. \mbox{$\omega_t, a_t \in\R^3$} are the true angular velocity and true linear acceleration, respectively, both expressed in the IMU frame. $n_t^g$ and $n_t^a$ are the Gaussian noise. $b_t^g, b_t^a \in \R^3$ are the biases of the gyroscope and accelerometer, respectively.  The bias is modeled as a Brownian motion as
\begin{equation}
    \dot b_t^g = \eta_g, \quad \dot b_t^a = \eta_a,
\end{equation}
where $\eta_g$ and $\eta_a$ follow the zero-mean Gaussian distribution. However, this simplified model can significantly deviate from the true behavior, resulting in inaccurate estimates of $w_t,a_t$.

The IMU measurements are often utilized with the following kinematics of a single rigid body \cite{hartley2020InEKF}:
\begin{equation}
    \begin{aligned}
        \dot R_t = R_t \omega_t^\times, \quad 
        \dot v_t = R_ta_t+g, \quad 
        \dot p_t = v_t,
    \end{aligned}
    \label{eq:1}
\end{equation}
where $R_t\in \mathrm{SO}(3)$ is the orientation of the body frame. $\mathrm{SO}(3)$ is the \emph{special orthogonal group}. $v_t\in\R^3$ is the linear velocity of the body frame expressed in the world frame, $p_t\in\R^3$ is the position of the body frame. $(\cdot)^\times:\R^3\to \mathfrak{so}(3)$ is a cross-product operator that satisfies $a^\times b=a\times b,\ \forall a,b\in\R^3$, where $\mathfrak{so}(3)$ is the vector space of $3$-by-$3$ \emph{skew-symmetric matrices} which is also the \emph{Lie algebra} of $\mathrm{SO}(3)$. $g\in\R^3$ is the gravity expressed in the world frame. In this work, we assume the IMU frame is the body frame.

Accurate pose (represented as a triple $(R,v,p)$ in the paper) is critical in robotics applications, which often involve integrating IMU measurements. However, direct integration of the measured \(\tilde{\omega}_t\) and \(\tilde{a}_t\) in \eqref{eq:1} leads to rapidly accumulating errors in pose over time due to noise and biases. Therefore, to get accurate integration results, it is necessary to develop a ``filter'' to give accurate or clean angular velocity and acceleration from raw IMU data.

For simplicity of notation, we will use the subscript~$x_k$ to represent the value of $x$ at time $t_k$, indicating the discretization of the continuous variable $x_t$ throughout the rest of the paper.

\subsection{Problem Formulation}
To describe the problem of getting the accurate IMU data from raw measurements, we can formulate it as follows:
\begin{prob}
\label{prob:1}
    Consider true $\omega_t$ and $a_t$ as deterministic parameters. Given the raw discrete measurement $\tilde{\omega}_{0:k}$ and $\tilde{a}_{0:k}$ from the initial time $t_0$ to current time $t_k$, find a suitable estimator for $\omega_k$ and $a_k$:
    \begin{equation}
        (\hat \omega_{k}, \hat a_{k}) = \mathcal F(\tilde{\omega}_{0:k},\tilde{a}_{0:k}),
    \end{equation}
    where $\mathcal F$ represents the estimator.
\end{prob}

We aim to develop a casual estimator that can operate online using only the \emph{past} IMU measurements during inference time. 
This is an open problem that remains to be explored and is also challenging due to several factors. First, even small errors in $w_t$ and $a_t$ can accumulate into significant pose errors due to the integration process, which requires the estimator to meet strict accuracy standards. 
Second, the estimator uses only IMU information rather than fusing it with other sensors. The information resource is limited, requiring us to carefully design a suitable IMU measurement model for the estimator.
Third, obtaining ground truth for $\omega_t$ and $a_t$ is often impractical, making it challenging to establish a suitable evaluation metric for the estimator. In addition, this makes data-driven approaches particularly difficult, as they depend on suitable loss and reliable ground truth for training.

\section{Learning Bias Dynamics}
In this section, we propose an estimator for problem \ref{prob:1} by using a neural ordinary differential equation framework~\cite{chen2018neuralode} for bias dynamics modeling.

\subsection{IMU Measurement Model}
Adopting~\eqref{eq:2}, we model the IMU measurements as corrupted by some biases and additive Gaussian noise.  Instead of modeling the bias dynamics as Brownian motions, we consider the bias $b_t^g$ and $b_t^a$ to be some deterministic variables with nonlinear dynamics as:
\begin{equation}
    \dot b_t^g = f_g(b_t^g, \tilde{u}_t), \quad \dot b_t^a = f_a(b_t^a, \tilde{u}_t),
    \label{eq:6}
\end{equation}
where $\tilde u_t:=(\tilde \omega_t,\tilde a_t)$ denotes the raw IMU measurements. Since $u_t$ serves as the control input to the bias dynamics, \eqref{eq:6} can be viewed as input-dependent vector fields. As a result, with known initial conditions, one can simply integrate the deterministic dynamics to obtain the bias at any time. 

With the known bias from \eqref{eq:6}, combining model \eqref{eq:2}, we can obtain an estimate of $w_t,a_t$ as 
\begin{equation}
    \hat \omega_t = \tilde{\omega}_t - b_t^g, \quad \hat a_t = \tilde{a}_t - b_t^a.
    \label{eq:5}
\end{equation}
This estimation can be regarded as that the noise is negligible. Such an estimator is reasonable because the estimation in \eqref{eq:5} corresponds to a Least-Squares Estimation (LSE) applied to the model $\tilde{\omega}_t - b_t^g = w_t + n_t^g$ using only a single measurement. A more comprehensive treatment of the noise term can be achieved by integrating additional sensor data within a multi-sensor fusion framework such as \cite{wisth2023VILENS,lin2023Drift}. In this work, however, we focus on the estimation method described in \eqref{eq:5}, leaving noise handling as the future work.

 According to \eqref{eq:6} and \eqref{eq:5}, estimating the true $w_t,a_t$ at time $t$ only requires IMU measurements and the initial condition. The initial bias condition can always be determined from stationary IMU data. Therefore, this provides a solution to problem \ref{prob:1}.
\begin{rmk}
    Compared to the \emph{stochastic differential equation} (SDE) model for bias in conventional assumption, our method can be regarded as only considering the \emph{drift} of the SDE, which reduces to an ordinary differential equation.
\end{rmk}

\subsection{Neural Ordinary Differential Equations}
\label{Sec:NODE}
We model the vector fields in~\eqref{eq:6} with a \emph{neural ordinary differential equation} (NODE) architecture introduced in \cite{chen2018neuralode}. The NODE seeks to model the dynamics of a state as $\dot z_t = f(z_t;\theta)$ using a neural network, where $\theta$ is the parameters of the network. Instead of directly fitting $\dot z_t$, the loss $L(\cdot)$ of the network is defined  based on $z_t$ through an ODE solver:
\begin{equation}
    L(z_T) = L(z_0 + \int_{t_0}^{t_T} f(z_t;\theta)\D t),
    \label{eq:7}
\end{equation}
where the integral can be solved by an ODE solver with methods such as the Euler method, RK45, etc. 
Therefore, the NODE framework takes the initial condition $z_0$ and time $t_T$, and outputs the state $z_T$ after integration.
To train this network, the gradient can be calculated by numerical backpropagation or memory-efficient \emph{adjoint} method~\cite{chen2018neuralode,kidger2021fasteradjoint}. Such a NODE framework allows us to use the sequence of observations of the state $\tilde z_{0},\tilde z_{1},...,\tilde z_{T}$ to fit the dynamics $\dot z_t$.


The ODE in~\eqref{eq:6} depends on additional control inputs $u_t$, which can not be handled by the canonical NODE framework. Moreover, the control inputs are measured in a discrete manner rather than a continuous formulation. To address this issue, we utilize the idea of \emph{neural controlled differential equations} \cite{kidger2020neuralcde}. The discrete control input $u_{i}\in\R^n$ can be interpolated using a continuous spline $\mathcal S:[t_0,t_T]\to \R^n$ such that $\mathcal S(t_i)=u_{i}$, where $t_0\leq t_i\leq t_N$. Therefore, the control input can be modeled as a function of $t$ given discrete $\tilde u_i$. To maintain consistency with the formulation in \eqref{eq:7}, where the dynamics do not explicitly depend on $t$, the time $t$ can be augmented to the original state as $z_{\text{aug}}=[z_t,t]$.

Define $b_t:=(b_t^g,b_t^a)$ and apply a continuous spline described above to handle the discrete control input $\tilde{u}_i$, we can reformulate the underlying bias dynamics \eqref{eq:6} in the NODE framework as
\begin{equation}
    \begin{bmatrix}
        \dot t \\
        \dot b_t
    \end{bmatrix} = \begin{bmatrix}
        1 \\ f(b_t,\mathcal{S}_{\tilde u}(t);\theta)
    \end{bmatrix}.
\end{equation}
Using this ODE, the bias at time $t_T$ is given by
\begin{equation}
    b_{T} = b_{0} + \int_{t_0}^{t_T} f(b_t,\mathcal{S}_{\tilde u}(t);\theta)\D t.
    \label{eq:9}
\end{equation}
$t$ can be ignored in the loss function since it does not depend on network parameters. After defining suitable networks and the loss function of $b_t$, we can use the NODE framework to learn the bias dynamics.

\section{Network Structure and Implementation}
\begin{figure*}
    \centering
    \includegraphics[width=0.9\textwidth]{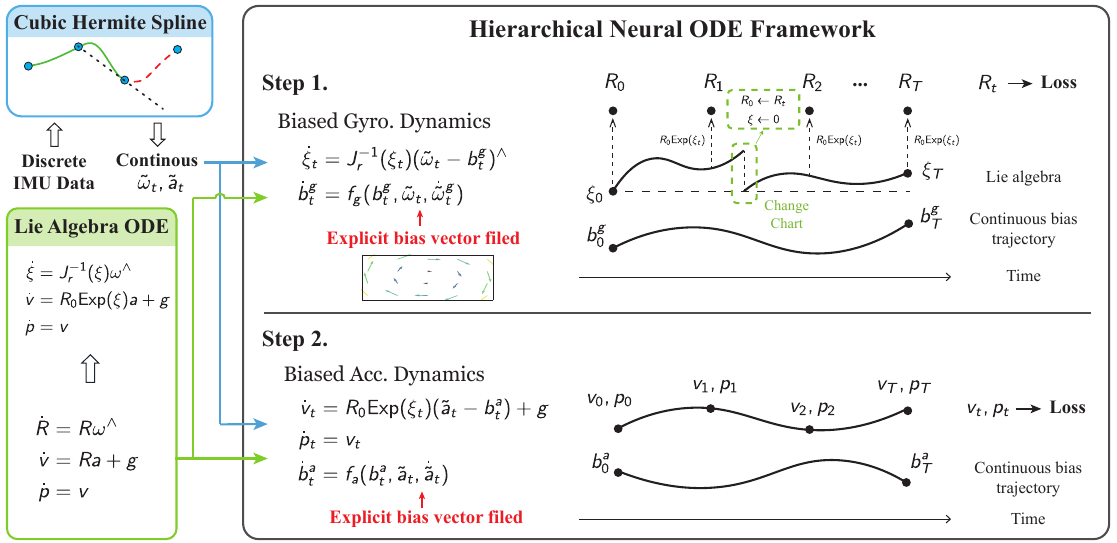}
    \caption{The framework for learning bias dynamics. 
    The bias dynamics are modeled by NODE, trained in a hierarchical manner. We first train the gyroscope component, followed by the accelerometer component. At each stage, the IMU raw data is represented as a continuous spline and serves as control input to the bias dynamics. Given initial conditions, the pose and bias along the trajectory are obtained through integration, with only the pose contributing to the loss function. The ODE on the manifold $\mathrm{SO}(3)$ is reformulated as a Lie algebra ODE, enabling an efficient solution via canonical NODE.
    }
    \label{fig:0}
    \vspace{-10px}
\end{figure*}

We have outlined the general concept of our proposed method; in this section, we present the detailed learning framework. The overall process is illustrated in Fig. \ref{fig:0}.

\subsection{Neural Network Structure and Input Spline}
\label{Sec:NeuralNetworkStructure}
To model the bias vector field, we utilize a simple multilayer perceptron with  residual connections \cite{he2016Resnet} for both gyroscope and accelerometer, which can be represented as 
\begin{equation}
    \dot b_t^g = \fgfunctheta, \quad \dot b_t^a = \fafunctheta,
    \label{eq:31}
\end{equation}
where $\theta$ represents the parameters of neural networks and is omitted in the rest of this paper for simplicity.
The explicit modeling of the bias vector field allows the network to remain lightweight, making it significantly simpler than most existing methods. This reduced model complexity lowers computational resource requirements. In our experiments, we find the residual connections can accelerate the model convergence.

In \eqref{eq:31}, we include additional derivatives $\dot {\tilde \omega}_t$ and $\dot{\tilde a}_t$ as the input, which is uncommon in the literature. This additional information allows us to model the components of bias that depend on $\omega_t$ and $a_t$. For example, if the measurement is linear in the true value as $\tilde \omega_t = A\omega_t$,
where $A\in\R^{3\times 3}$ is invertible, the bias dynamics should be
\begin{equation}
    \dot b_t^g = \frac{\D}{\D t}(\tilde{\omega}_t-\omega_t) = (I-A^{-1})\dot{\tilde{\omega}}_t,
\end{equation}
which is a function of $\dot {\tilde{\omega}}_t$. This demonstrates the importance of including $\dot {\tilde{\omega}}_t$ and $\dot {\tilde{a}}_t$, as the linear model is often used to model axis misalignment in an IMU~\cite{rehder2016kalibr, brossard2020denoising}.


To handle the input into a continuous function of time $t$ as \eqref{eq:9}, we utilize a \emph{cubic Hermite spline} with the rule of backward differences \cite{morrill2022choicecdespine} to interpolate the discrete input. This is a cubic spline $\mathcal S_{\tilde u}$ satisfying the following:
\begin{equation}
    \mathcal S_t(t_i) = \tilde u_{i}, \quad \dot{\mathcal S}_t(t_i) = (\tilde u_{i} - \tilde u_{i-1})/ (t_i-t_{i-1}),
    \label{eq:spline}
\end{equation}
where $(t_i,\tilde u_{i})$ is the discrete data. Therefore, the continuous spline for IMU measurements is given by $(\tilde \omega_t,\tilde a_t)=\mathcal S_{\tilde u}(t)$. 
Such a spline has continuous derivatives, which can benefit the convergence of the NODE \cite{morrill2022choicecdespine}. More importantly, this spline is causal, which means the spline up to $t_i$ will not depend on $\tilde{u}_{i+1}$. This enables the spline to be generated in real-time as new measurements are received, allowing the network to perform online inference. 

\subsection{Loss Construction}
To train the NODE \eqref{eq:9}, a straightforward loss could be defined as a scalar function of estimated bias and the true bias $L(\hat b_t, b_t)$. However, obtaining the true bias is challenging. One can use multi-sensor fusion to estimate bias \cite{buchanan2022deepbias}, but this requires extra resources and depends on the accuracy of the estimation algorithm. We proposed directly using the pose ground truth to construct the loss for learning bias dynamics. 

Combining the bias dynamics \eqref{eq:6} and the IMU kinematics \eqref{eq:1}, we can get the ODE as
\begin{equation}
    \begin{aligned}
        \dot R_t &= R_t(\tilde{\omega}_t-b_t^g)^\times\\
        \dot b_t^g &= \fgfunc\\
        \dot v_t &= R_t(\tilde{a}_t-b_t^a)+g\\
        \dot p_t &= v_t\\
        \dot b_t^a &= \fafunc.
    \end{aligned}
    \label{eq:10}
\end{equation}
By solving the entire ODE, we can construct the loss function using the pose ground truth. For orientation, we apply a mean-squared error (MSE) loss defined as
\begin{equation}
    L_{R} = \frac{1}{N}\sum_{k=1}^{N}\|\Log(\hat R_k R_k^{\mathsf{T}})\|_2^2,
    \label{eq:16}
\end{equation}
where $\Log(\cdot):\mathrm{SO}(3)\to\R^3$ is the vectorized \emph{logarithm} of $\mathrm{SO}(3)$ \cite{forster2017onmanifold}, $\hat R_k$ is the solution by solving \eqref{eq:10}, $R_k$ is the ground truth. For velocity and position, we also use an MSE loss:
\begin{equation}
    L_{v,p}= \frac{1}{N}\sum_{k=1}^{N}(\|\hat{p}_k-p_k\|_2^2 + \|\hat{v}_k-v_k\|_2^2).
    \label{eq:17}
\end{equation}
where $\hat p_k,\hat v_k$ are the solution form \eqref{eq:10} and $p_k,v_k$ are the ground truth. Therefore, the total loss for training NODE \eqref{eq:10} is given by 
\begin{equation}
    L=L_R+L_{v,p}.
    \label{eq:loss_total}
\end{equation} 
By coupling the bias dynamics and the IMU kinematics in \eqref{eq:10}, the neural networks' parameters of bias dynamics can be optimized using the loss \eqref{eq:loss_total} that only depend on pose ground truth. 

\subsection{Solving ODE on Lie Group via Lie Algebra}
Embedding the ODE \eqref{eq:10} within the standard NODE framework presents a challenge since it involves the manifold $\mathrm{SO}(3)$, whereas the standard NODE framework is designed for Euclidean spaces. To address this, we utilize a Lie group method \cite{hairer2006GeometricNumericalIntegration}, reformulating the ODE on the \emph{Lie algebra}, a linear space naturally compatible with the NODE framework. In this section, we focus on addressing the first equation in \eqref{eq:10}, the non-Euclidean part, which can be easily integrated into the original ODE (see Sec. \ref{sec:5-4}). 
Consider the following differential equation:
\begin{equation}
    \dot R_t = R_t\omega_t^\times,
    \label{eq:14}
\end{equation}
where $\omega_t$ is a function that only depends on $t$.
The following theorem holds:
\begin{thm}
    \label{thm:1}
    The solution of the differential equation \eqref{eq:14} with initial condition $R(0)=R_0$ is given by $R_t=R_0\Exp(\xi_t)$, where $\xi_t\in\R^3$ is the solution of 
    \begin{equation}
        \dot \xi_t = J_r^{-1}(\xi_t)\omega_t, \quad \xi_t(0) = \bm{0},
        \label{eq:15}
    \end{equation}
    as long as $\|\xi_t\|<2\pi$. $J_r^{-1}\in\R^{3\times 3}$ is the inverse of the right Jacobian of $\mathrm{SO}(3)$~\cite{barfoot2017state,chirikjian2012stochastic}, which is well-defined under the condition $\|\xi_t\|<2\pi$. \hfill $\square$
\end{thm}

The proof is given in the Appendix. The proof generally follows \cite[Theorem 7.1]{hairer2006GeometricNumericalIntegration}, however, which gives a left derivative version for the general Lie groups. Here we provide the right derivative version with matrix formulation, which is more compatible with IMU kinematics.

The theorem indicates we can obtain the local solution of ODE \eqref{eq:14} by solving an ODE in Euclidean space. To solve the ODE on $\mathrm{SO}(3)$ over time, we need to change the \emph{chart} by resetting $R_n$ as the initial condition and $\xi_t$ to zeros before the Jacobian approaches the singularity, i.e., $\|\xi_t\|\to 2\pi$. 
The general idea can be summarized as follows: parameterize the point $R_0$ on $\mathrm{SO}(3)$ into $\R^3$ using the local chart around $R_0$, solving the corresponding ODE in $\R^3$ to determine the next desired point, and then map the point in $\R^3$ back to $\mathrm{SO}(3)$ to get $R_1$. Repeat this process then we can get the solution $R_t$, see Algorithm \ref{alg:1}. 
\begin{algorithm}[ht]
\caption{Solving ODE for rotation kinematics}
\label{alg:1}
\begin{algorithmic}[1]
\STATE \textbf{Input:} Initial condition $R_0$, input $\omega_t$, time $[t_0,t_{T}]$
\STATE Initialize $R_0 \gets R_0$, $\xi_t = 0$, $t\gets t_0$
\WHILE{$t\leq t_{T}$}
    \STATE $\xi_t=\mathtt{ODESolver}\{\dot \xi_t = J_r^{-1}(\xi_t)\omega_t\}$
    \STATE $R_t = R_0\Exp(\xi_t)$
    \STATE \textbf{if} $\|\xi_t\|>\pi$ \textbf{then} $R_0\gets R_t$, $\xi_t \gets 0$
    \STATE Increase $t$ (adaptive or fixed step, $t={t_0,t_1,...,t_T}$)
\ENDWHILE
\STATE \textbf{Output:} Solution $R_t$, where $t={t_0,t_1,...,t_T}$.
\end{algorithmic}
\end{algorithm}

Compared to the canonical Euclidean ODE, our approach introduces only an additional chart transition as $\xi_t$ approaches the singularity. We present a PyTorch-based ODE solver that extends the standard NODE framework to accommodate $\mathrm{SO}(3)$ dynamics. The gradient of this NODE is computed using numerical backpropagation. The corresponding adjoint method for calculating gradient is beyond the scope of this work, which can be the future work.


\begin{rmk}
The singularity issue in Theorem \ref{thm:1} typically requires changing the chart at each integration step to ensure well-definedness. However, in our application, the angular velocity will not jump too much. Therefore, we relax this condition and only change the chart when $\|\xi\|$ closes to $\pi$ to improve the computation efficiency. If singularities still occur, the threshold can be reduced further until chart updates are triggered at each step.
\end{rmk}

\subsection{Hierarchical Training Framework}
\label{sec:5-4}
The NODE framework for learning bias dynamics can be regarded as a continuous recurrent neural network (RNN) \cite{rubanova2019irregular_ODERNN}, which is hard to train. To address this, we propose a hierarchical strategy to make neural networks converge faster. 

Under our hypothesis, the biases of the gyroscope $b_t^g$ and accelerometer $b_t^a$, are decoupled. This means their dynamics are independent of each other. We can adopt a two-stage training process by leveraging the IMU kinematics in \eqref{eq:1}, which indicates that velocity and position are dependent on rotation but not vice versa. First, we train $\dot b_t^g$ for the rotational component, and once complete, we fix $\dot b_t^g$ to train $\dot b_t^a$.

For the rotational component, we use the method described in the previous section and solve the following ODE:
\begin{equation}
    \begin{aligned}
        \dot \xi_t &= J_r^{-1}(\xi_t)(\tilde{\omega}_t-b_t^g)\\
        \dot b_t^g &= \fgfunc,
    \end{aligned}
    \label{eq:20}
\end{equation}
where $R_t$ can be recovered by $R_t=R_0\Exp(\xi_t)$. The loss function $L_R$ is defined in \eqref{eq:16}. Once this stage is complete, the network parameters of $f_g$ are fixed, and we proceed by combining \eqref{eq:20} with the following equations to handle the accelerometer component:
\begin{equation}
    \begin{aligned}
        \dot v_t &= R_0\Exp(\xi_t)(\tilde{a}_t-b_t^a)+g\\
        \dot p_t &= v_t\\
        \dot b_t^a &= \fafunc,
    \end{aligned}
    \label{eq:21}
\end{equation}
where $R_t=R_0\Exp(\xi_t)$ has been substituted. The corresponding loss function $L_{v,p}$ is defined as \eqref{eq:17}. The entire training process is illustrated in Fig. \ref{fig:0}.

In each training stage, the most straightforward approach would be to train the neural ODE by integrating over the entire sequence, calculating the loss, and optimizing the parameters. However, this method is computationally expensive and memory-intensive, making it impractical for implementation. To address these limitations, we adopt the approach proposed in \cite{chen2018neuralode}, where training is performed using short segments of the sequence. Instead of integrating over the entire sequence, we compute the vector field by integrating over multiple short time intervals, significantly reducing computational cost and memory usage. This approach can be expressed as: 
\begin{equation}
    (\hat R, \hat v, \hat p)_{s:s+N} = \texttt{ODEINT}_{\mathrm{SO(3)}}(R_{s},b^g_s,v_s,p_s,b_s^a),
    \label{eq:30}
\end{equation}
where $s$ represents the initial time step of short time intervals and $N$ represents the length of the intervals. In the implementation, we set $N=16$, making a trade-off between being sufficiently long to effectively learn the vector field and short enough to avoid excessive training time. This training strategy requires the initial condition for each interval. Usually, we can get the pose ground truth but hard to the biased ground truth. We utilize an approximation of the initial bias condition as
\begin{equation}
    \begin{aligned}
        b_k^g &= \tilde{\omega}_k - \Log(R_k^TR_{k+1})/(t_{k+1}-t_k)\\
        b_k^a &= \tilde{a}_k - R_k^T((v_{k+1}-v_{k})/(t_{k+1}-t_k)-g),
    \end{aligned}
\end{equation}
where $R_k$, $v_k$ is assumed known as the ground truth. 
The initial bias for each time interval, determined in this manner, is affected by noise from IMU measurements and pose estimates. However, as long as the noise remains within a reasonable range, the proposed method can effectively manage this uncertainty. In this context, the noise in the bias initialization can be regarded as data augmentation.

\section{Experimental Results}
In this section, we demonstrate the proposed method's accuracy in angular velocity and acceleration. We compare pure IMU integration and its application in visual-inertial odometry against other methods on two public datasets. We also evaluate its generalization capability on a real-world experiment. 

\subsection{Public Datasets and Training Platform}

\subsubsection{\texorpdfstring{EUROC \cite{burri2016euroc}}{}} 
This is a VIO dataset for a micro aerial vehicle (MAV). It comprises 11 trajectories with lengths ranging from approximately 36.5 to 130.9 meters and durations between 99 to 182 seconds. The IMU used is the ADIS16448, operating at 200 Hz, providing the angular velocity and the linear acceleration. The dataset includes stereo camera data at 20 Hz. Ground truth poses are provided by a motion capture system and a laser tracker.

\subsubsection{\texorpdfstring{TUM-VI \cite{schubert2018tum}}{}} This is a handheld VIO dataset. It contains 28 sequences, covering a total distance of approximately 20 km, with both indoor and outdoor scenes. The IMU used is the Bosch BMI160, which also provides the angular velocity and the linear acceleration at a frequency of 200 Hz. The dataset contains the stereo camera data at 20 Hz. Ground truth poses are obtained from a motion capture system but are available only for the indoor environment, resulting in some sequences with incomplete pose ground truth. For our experiments, we select the 6 room sequences, which contain the longest ground truth data, each lasting 2–3 minutes.

\subsubsection{Data Arrangement and Platform}
We divide the data into training, validation, and test sets for both the EUROC and TUM-VI datasets. For training and validation, we use the same sequences, with the first 80\% designated for training and the last 20\% for validation. The test set consists of entirely unseen sequences. In the EUROC dataset, following \cite{brossard2020denoising}, we use 6 sequences for training and the remaining 5 for testing. For the TUM-VI dataset, 3 sequences are used for training and 3 for testing.

For both datasets, the data is assumed to be synthesized, with raw IMU data serving as the network’s input. Different from EUROC, TUM-VI also provided model-based calibrated IMU data. The training is processed on a laptop with RTX4060, 8G memory. We choose the Adam-optimizer \cite{kingma2014adamoptimizer} with a StepLR scheduler for the learning rate. The epoch is chosen 1800 for all trainings. For the EUROC dataset, the learning rate is set as 0.005 for both the gyroscope and accelerometer. We set nearly the same parameters for TUM-VI dataset. 
The integration method chosen is the Euler method for fast training.

\subsection{Evaluation Metric}
We utilize the \emph{absolute error} and \emph{relative error} \cite{zhang2018tutorialATE} as the metric, which are the common criteria in VIO systems. More specifically, the following metric in toolbox \emph{evo} \cite{grupp2017evo} is used.
\subsubsection{Absolute Error}
After the alignment of the estimated trajectory and ground trajectory, the \emph{Absolute Orientation Error} (AOE) is defined as
\begin{equation}
    \text{AOE} = \sqrt{\frac{1}{N}\sum_{k=1}^{N}\|\Log(\hat R_k^{\mathsf T} R_k)\|^2_2},
\end{equation}
where $\hat R_k$ is the estimated orientation and $R_k$ is the ground truth. The \emph{Absolute Position Error} (APE) is defined as
\begin{equation}
    \text{APE} = \sqrt{\frac{1}{N}\sum_{k=1}^{N}\|p_k-\hat p_k\|^2_2},
\end{equation}
where $\hat p_k$ is the estimated position and $p_k$ is the ground truth.

\subsubsection{Relative Error}
\begin{figure}
    \centering
    \includegraphics[width=\linewidth]{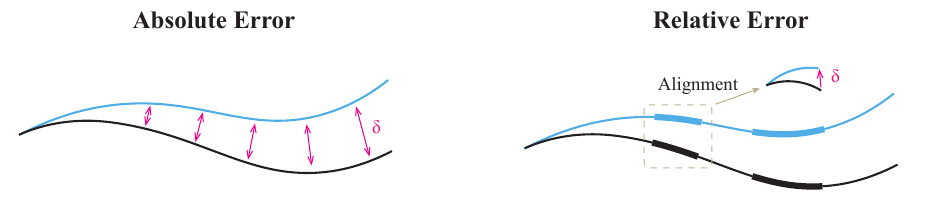}
    \caption{Absolute error and relative error.}
    \label{fig:relativeerror}
\end{figure}

The idea of relative error is to compare multiple sub-trajectories of two trajectories rather than the entire ones. To define the relative error, all sub-trajectories whose length is $d$~m will be collected using the ground truth, denote $X_{s_0:s_1}$, where $X\in SE(3)$ made of $R$ and $p$. The increment for each sub-trajectory will be calculated using
$\Delta X_s = X_{s_0}^{-1}X_{s_1}$. Then the difference between the estimation and ground truth for each sub-trajectory is defined as $X_{e,s}=\Delta X_s^{-1} \Delta \hat X_s$. Extracting the orientation and position components, $X_{e,s} = (R_{e,s},p_{e,s})$, we obtain the \emph{Relative Orientation Error} (ROE) and \emph{Relative Position Error} (RPE) as follows:
\begin{equation}
    \text{ROE} = \frac{1}{D}\sum_{s=1}^{D}\|\Log(R_{e,s})\|_2\quad \text{RPE} = \frac{1}{D}\sum_{s=1}^{D}\|p_{e,s}\|_2.
\end{equation}
Fig. \ref{fig:relativeerror} shows the difference between absolute errors and relative errors. Since absolute error is sensitive to the time that errors occur \cite{zhang2018tutorialATE,geneva2020openvins}, relative error provides more informative insights into the comparison of two trajectories.

\subsection{Compared Methods}
In each experiment, the IMU data from the following sources are used.
\begin{enumerate}
    \item \textbf{Raw IMU}: Raw uncalibrated data.
    \item \textbf{Linear}: A simple linear model with constant biases, $\hat \omega = A_g\tilde{\omega}+b_g$, $\hat a = A_a\tilde{a}+b_a$. The parameters $A$ and $b$ are modeled using a single linear layer network, which is trained by considering $\dot b\equiv 0$ in the proposed framework.
    \item \textbf{M.B.} \cite{brossard2020denoising}: A CNN-based neural network, which implicitly inference the bias using a window of IMU raw data. However, it only provides the Gyro. calibration. The parameters for their method are set to the defaults provided in their open-source project.
    \item \textbf{Proposed}: The calibrated IMU data by the proposed method.
\end{enumerate}

\subsection{Debiasing Results on EUROC and TUM-VI}
To demonstrate that the proposed method provides more accurate angular velocity and acceleration estimates, we present two types of results. First, we integrate the IMU measurements alone and evaluate the orientation and position errors. Second, we substitute the debiased IMU data into a VIO algorithm and examine the resulting orientation and position errors. 
The brief IMU grades are shown in Tab. \ref{tab:IMUgrade} for a more intuitive understanding of the improvements of the proposed method.



\begin{figure}
    \centering
    \includegraphics[width=\linewidth]{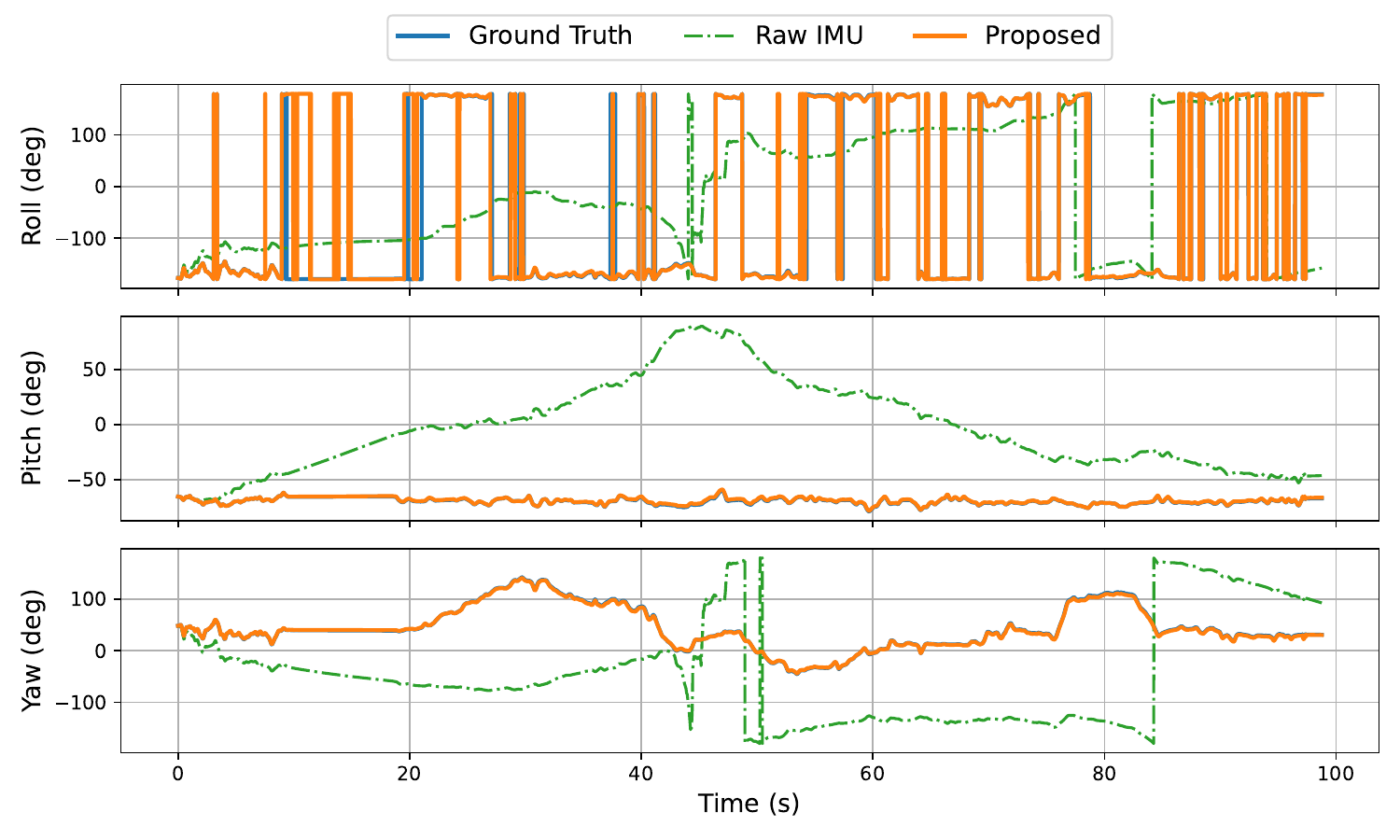}
    \caption{The Euler angles obtained by only integrating the IMU data. The results are very close to the ground truth, which implies that the debiased gyroscope yields good performance.}
    \label{fig:singleGyro}
    \vspace{-10px}
\end{figure}
\begin{figure}[ht]
    \centering
    \subfloat[EUROC: MH\_04. Pure integration orientation errors.\label{fig:1a}]{\includegraphics[width=\linewidth]{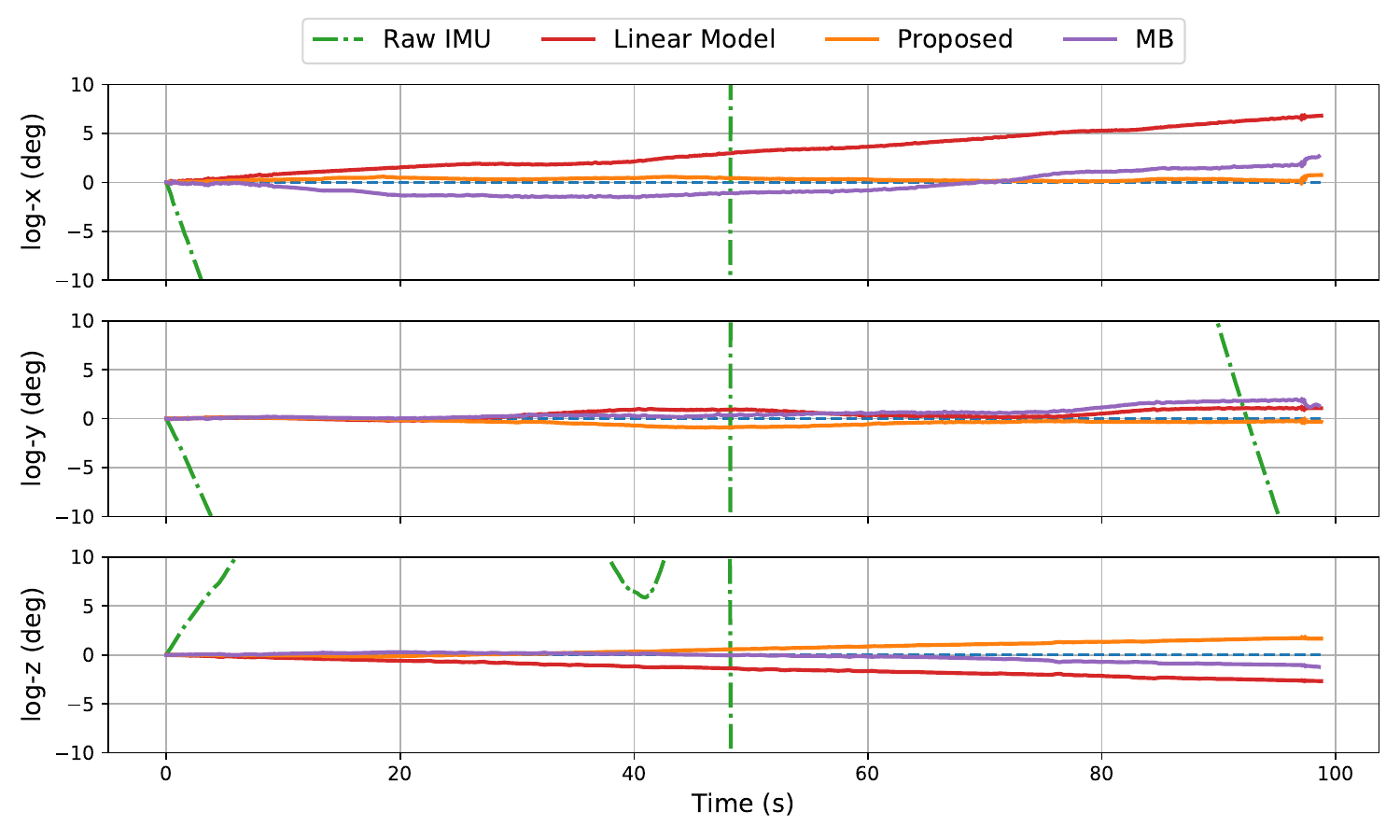}}
    \hfill
    \subfloat[TUM: Room4. Pure integration orientation errors.\label{fig:1b}]{\includegraphics[width=\linewidth]{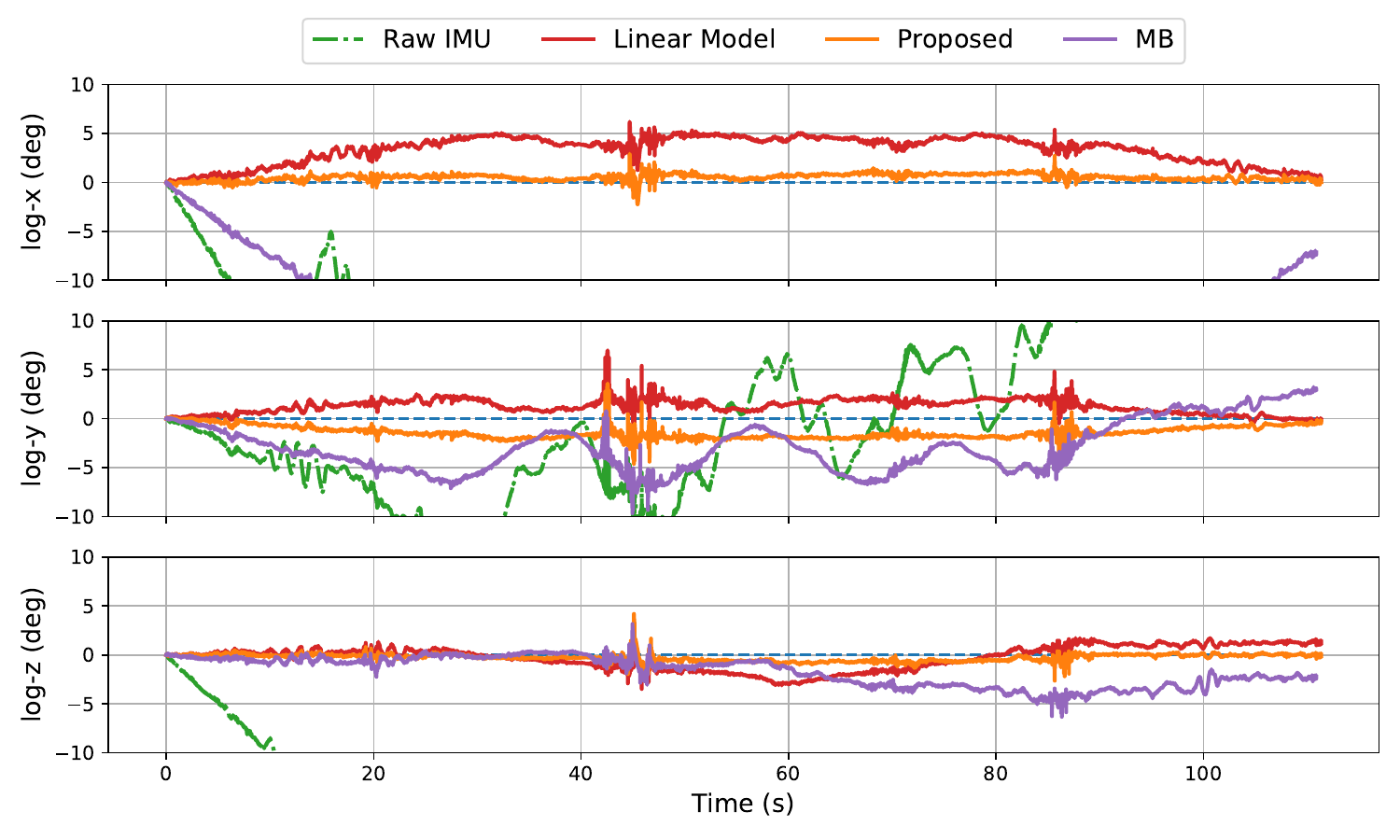}}
    \hfill
    \subfloat[FETCH: 05\_random. Pure integration  orientation errors.\label{fig:1c}]{\includegraphics[width=\linewidth]{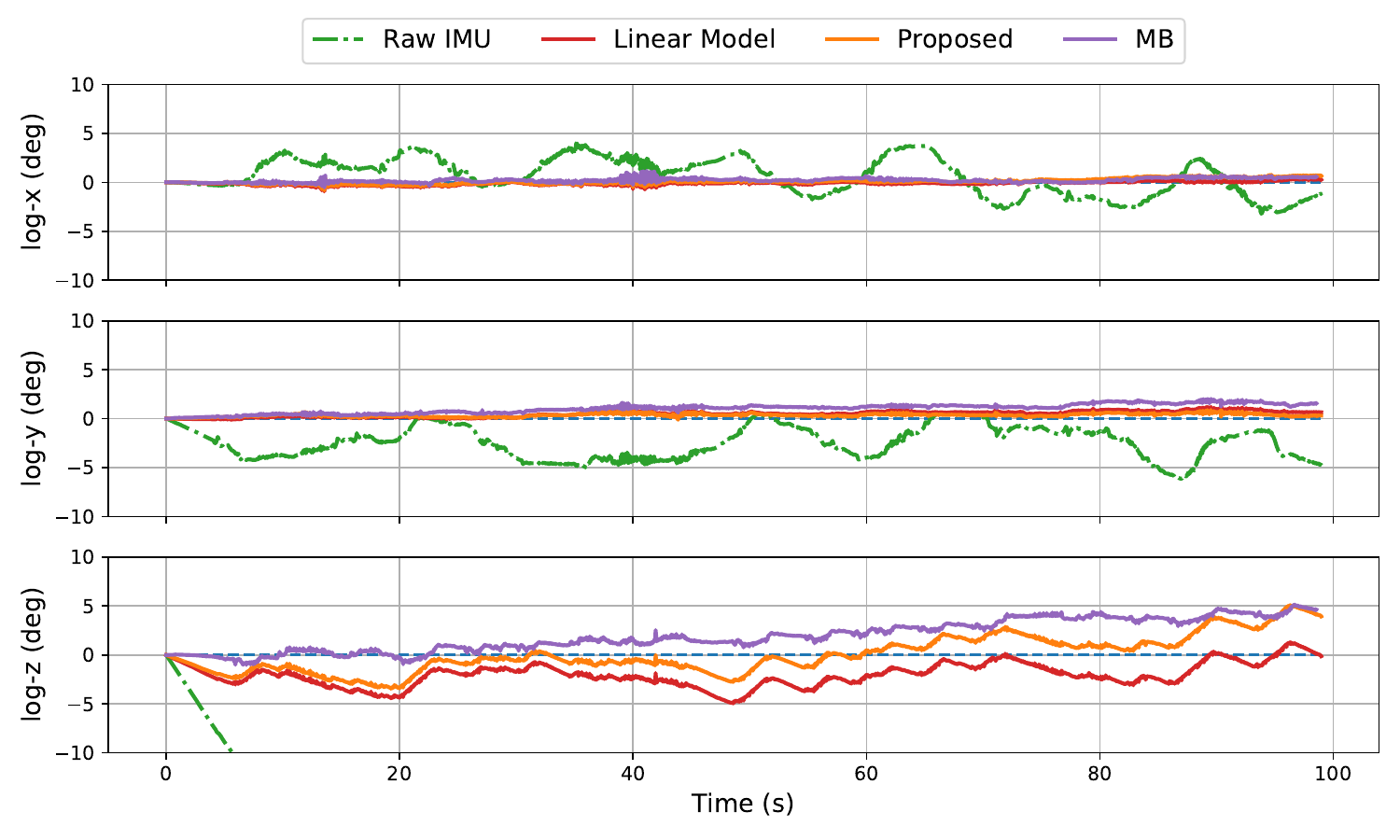}}
    \caption{The coordinates of the errors, $\Log(R_{gt}^\mathsf{T}\hat R)$. The estimates are obtained by only integrating the IMU data. The unit is converted to a degree. Closer to zero yields better results.}
    \label{fig:1}
    \vspace{-10px}
\end{figure}
\begin{figure}[ht]
    \centering
    \subfloat[EUROC: MH\_04. Pure integration velocity estimation. \label{fig:2a}]{\includegraphics[width=\linewidth]{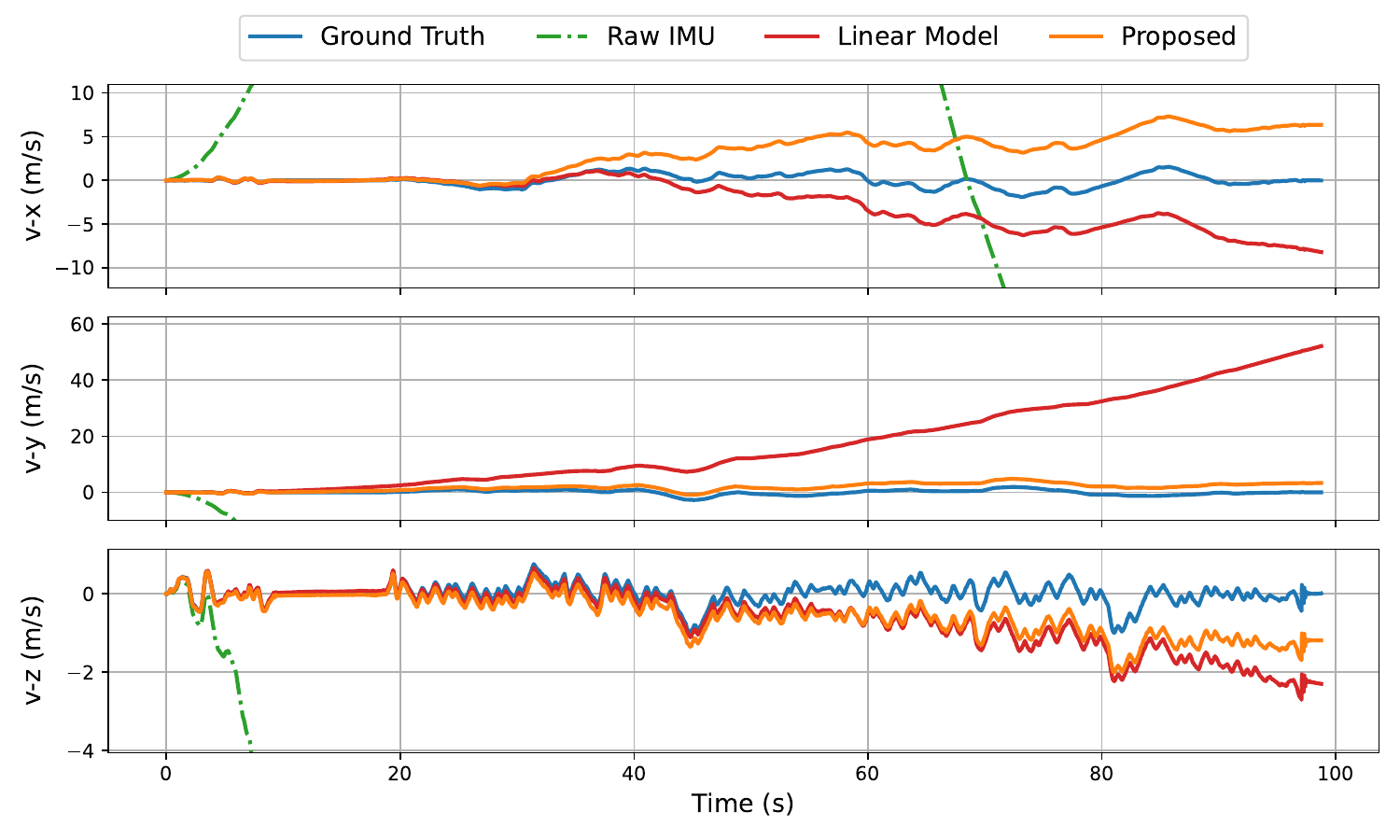}}
    \hfill
    \subfloat[TUM: Room4. Pure integration velocity estimation.\label{fig:2b}]{\includegraphics[width=\linewidth]{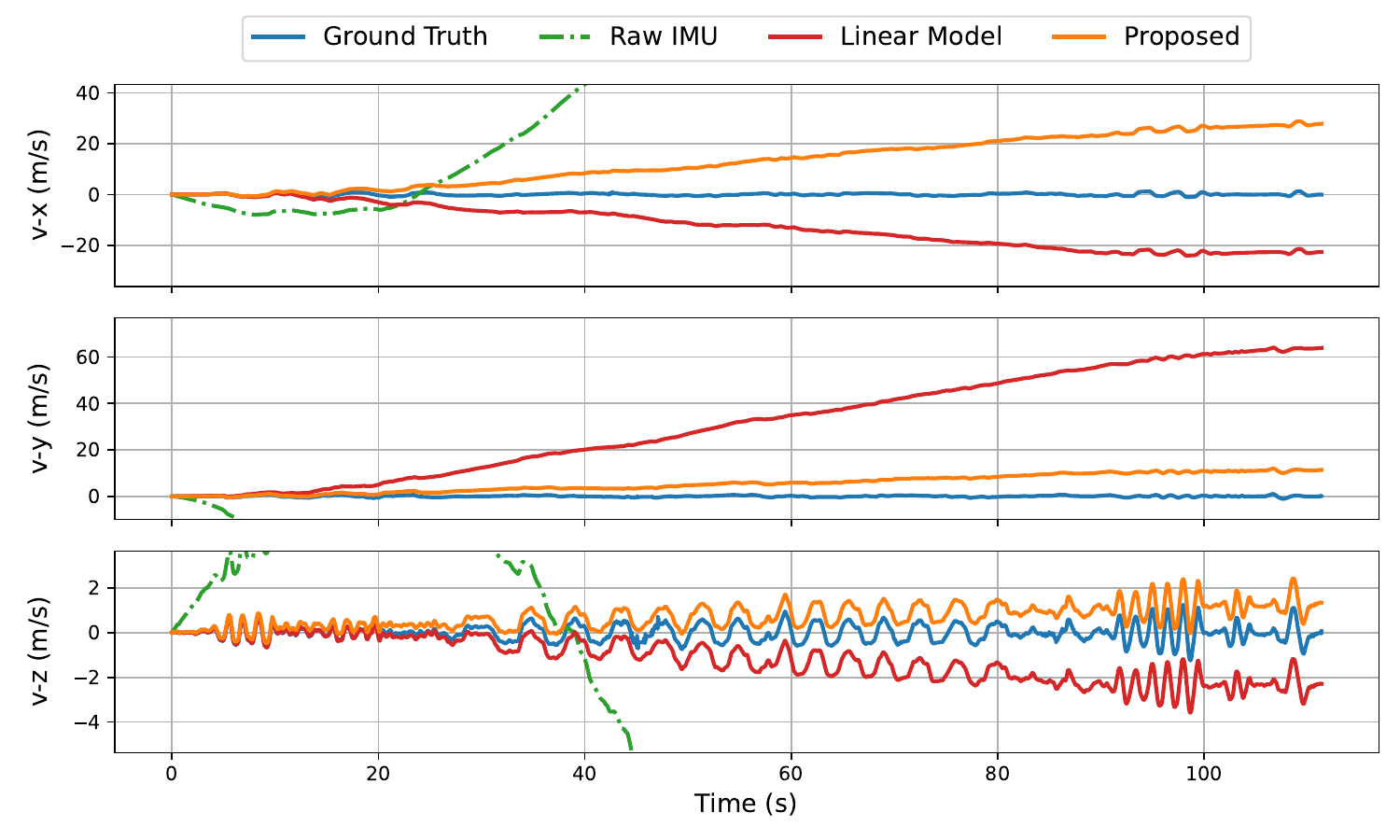}}
    \hfill
    \subfloat[FETCH: 05\_random. Pure integration velocity estimation.\label{fig:2c}]{\includegraphics[width=\linewidth]{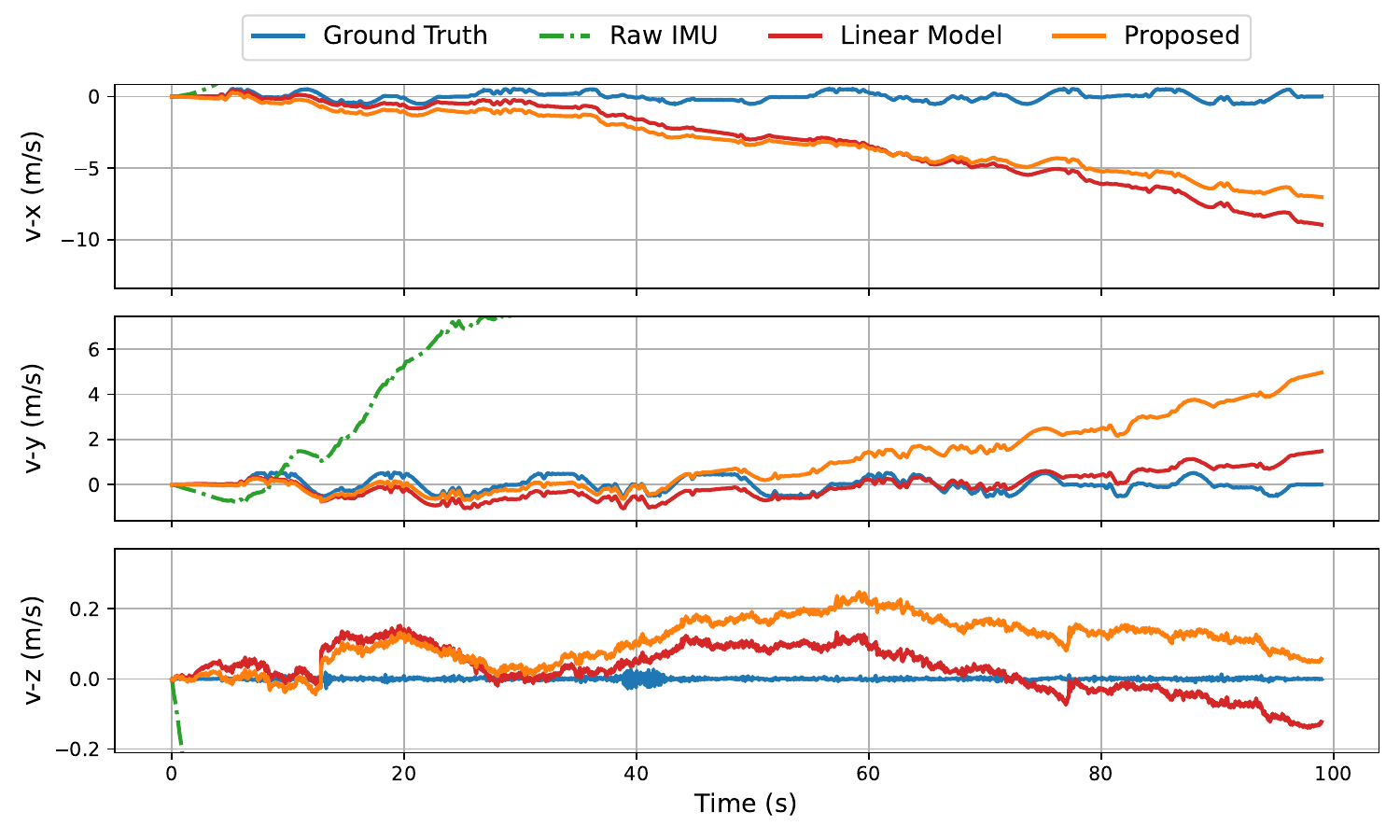}}
    \caption{The Velocity estimation obtained by only integrating the IMU data. Closer to the ground truth yields better results.}
    \label{fig:2}
    \vspace{-10px}
\end{figure}

The absolute and relative pose errors from pure IMU integration are shown in Table \ref{tab:5} and \ref{tab:6}, respectively. Since the method in \cite{brossard2020denoising} only provides angular velocity estimates, raw acceleration data is used. This scenario reflects cases where camera data is unavailable, only IMU provides odometry information for a VIO. It can be found that raw IMU data is unreliable, implying it is necessary to have a calibration for IMU. The proposed method outperforms other approaches in both absolute and relative error metrics. Notably, orientation errors are significantly smaller than position errors. Our findings show that pure integration with debiased IMU data yields accurate orientation along the test trajectories, see Fig.~\ref{fig:singleGyro}. The comparison of different method for orientation is shown in Fig.~\ref{fig:1a} and \ref{fig:1b}, which only shows the error of estimations for better visualization. Calibrated IMU generally gives good results and the proposed method gives better results than other compared methods. 
However,  position estimates remain unreliable due to the inherent challenges of double integration and sensitivity to orientation estimation. Instead, we present the velocity estimation in Fig. \ref{fig:2a} and \ref{fig:2b}. The result from the M.B. method \cite{brossard2020denoising} is omitted, as it does not account for accelerometer debiasing. The velocity estimation is reliable only over short intervals, while long-term integration leads to divergence. The significant velocity error is partly due to its dependence not only on accelerometer measurements but also on the accuracy of orientation estimation. The velocity along the z-axis (yaw axis) exhibits the smallest error. A possible reason is that z-axis velocity is influenced by pitch and roll, which tend to be less aggressive than yaw during most motions.
Although pure integration over long trajectories is rare in practice, these results highlight the strengths of the proposed method.

\begin{table*}[ht]
    \renewcommand{\arraystretch}{1.1}
    \footnotesize
    \centering
    \caption{Absolute Orientation Error (AOE) and Absolute Position Error (APE) of pure IMU integration and a VIO for EUROC and TUM-VI. Less is better. The pure integration results in position from raw IMU data have been removed due to significant errors.
    }
    \begin{tabular}{cccccccccccc}
         \hline

        \hline
        \multirow{2}{*}{\textbf{Dataset}} &\multirow{2}{*}{\textbf{Seq.}}& &\multicolumn{4}{c}{\textbf{AOE / APE of IMU Integration (deg / m)}} & &\multicolumn{4}{c}{\textbf{AOE / APE of VIO (deg / m)}} \\
         & & &\textbf{Raw IMU} & \textbf{Linear} & \textbf{M.B. \cite{brossard2020denoising}} & \textbf{Proposed} &&\textbf{Raw} & \textbf{Linear} & \textbf{M.B. \cite{brossard2020denoising}} & \textbf{Proposed}\\
        \hline
        \multirow{6}{*}{\textbf{EUROC}} 
& MH\_02 & & 123.05/- & 5.01/2204.16 & 4.54/1647.48 & \textbf{3.12}/\textbf{249.79} & & 1.73/0.30 & \textbf{0.86}/\textbf{0.19} &  1.41/0.21 & 1.01/0.22 \\
& MH\_04 & & 130.33/- & 4.08/636.25 & 1.47/433.94 & \textbf{1.06}/\textbf{145.40} & & 2.07/0.43 & 0.95/0.82 &  1.52/1.16 & \textbf{0.65}/\textbf{0.31} \\
& V1\_01 & & 113.47/- & 4.89/2031.47 & 1.93/1862.44 & \textbf{1.27}/\textbf{1712.75} & & 1.83/0.12 & 1.83/0.11 &  \textbf{1.20}/\textbf{0.10} & 1.44/0.11 \\
& V1\_03 & & 120.13/- & 2.52/693.59 & \textbf{1.58}/\textbf{407.72} & 2.26/614.36 & & 2.82/\textbf{0.11} & 1.16/0.11 &  1.67/0.16 & \textbf{0.85}/0.12 \\
& V2\_02 & & 116.92/- & 4.62/1210.81 & 4.85/\textbf{494.58} & \textbf{4.27}/1111.75 & & \textbf{1.87}/\textbf{0.15} & 2.26/0.16 &  2.16/0.16 & 2.60/0.16 \\
& \textbf{Average} & & 120.78/- & 4.22/1355.26 & 2.87/969.23 & \textbf{2.40}/\textbf{766.81} & & 2.07/0.22 & 1.41/0.28 & 1.59/0.36 & \textbf{1.31}/\textbf{0.18} \\
         \hline
         \multirow{4}{*}{\textbf{TUM-VI}}
         & Room2 & & 32.82/- & 15.02/3355.75 & 15.60/12268.32 & \textbf{1.69}/\textbf{1216.82} & & failed & 10.04/0.39 &  9.80/\textbf{0.32} & \textbf{5.08}/0.35 \\
& Room4 & & 70.19/- & 4.07/1615.27 & 17.90/7782.74 & \textbf{1.71}/\textbf{672.13} & & failed & 2.56/0.22 &  9.58/0.23 & \textbf{2.52}/\textbf{0.11} \\
& Room6 & & 65.15/- & 8.47/2333.58 & 19.10/11669.17 & \textbf{1.97}/\textbf{1139.77} & & failed & \textbf{1.55}/0.10 &  8.92/0.15 & 3.04/\textbf{0.05} \\
& \textbf{Average} & & 56.06/- & 9.19/2434.87 & 17.53/10573.41 & \textbf{1.79}/\textbf{1009.58} & & failed & 4.72/0.24 & 9.43/0.23 & \textbf{3.55}/\textbf{0.17} \\

        \hline
    \end{tabular}
    \label{tab:5}
\end{table*}
\begin{table*}[ht]
    \renewcommand{\arraystretch}{1.1}
    \centering
    \caption{Relative Orientation Error (ROE) and Relative Position Error (RPE) of pure IMU integration and a VIO for EUROC and TUM-VI. Less is better. Len. represents the given distance for finding evaluation pairs.}
    \begin{tabular}{cccccccccccc}
         \hline

        \hline
        \multirow{2}{*}{\textbf{Dataset}} &\multirow{2}{*}{\textbf{Len. (m)}}& &\multicolumn{4}{c}{\textbf{ROE / RPE of IMU Integration (deg / m)}} & &\multicolumn{4}{c}{\textbf{ROE / RPE of VIO (deg / m)}} \\
         & & &\textbf{Raw IMU} & \textbf{Linear} & \textbf{M.B. \cite{brossard2020denoising}} & \textbf{Proposed} &&\textbf{Raw} & \textbf{Linear} &\textbf{M.B. \cite{brossard2020denoising}} & \textbf{Proposed}\\
        \hline
        \multirow{5}{*}{\textbf{EUROC}} 
         & 5 & & 38.51/4823.72 & 0.99/241.61 & 0.89/167.74 & \textbf{0.71}/\textbf{133.79} & & 0.71/0.09 & 0.67/0.11 & 0.73/0.11 & \textbf{0.65}/\textbf{0.08} \\
& 10 & & 62.11/9290.19 & 1.44/456.60 & 1.25/321.93 & \textbf{0.98}/\textbf{256.06} & & 0.93/0.12 & 0.93/0.16 & 0.96/0.15 & \textbf{0.88}/\textbf{0.11} \\
& 15 & & 79.74/14029.04 & 1.81/690.52 & 1.47/487.35 & \textbf{1.21}/\textbf{390.69} & & 1.01/0.14 & 1.04/0.19 & 1.04/0.19 & \textbf{0.97}/\textbf{0.14} \\
& 20 & & 93.17/18735.49 & 2.18/919.75 & 1.71/654.11 & \textbf{1.44}/\textbf{517.72} & & 1.08/0.16 & 1.16/0.22 & 1.14/0.23 & \textbf{1.06}/\textbf{0.15} \\
         \hline
         \multirow{5}{*}{\textbf{TUM-VI}}
          & 5 & & 12.12/2546.67 & 1.58/356.56 & 4.47/1604.39 & \textbf{0.85}/\textbf{156.37} & & failed & 1.07/0.07 & 1.54/0.06 & \textbf{0.86}/\textbf{0.03} \\
& 10 & & 17.98/4829.26 & 2.45/685.14 & 6.26/3083.10 & \textbf{1.00}/\textbf{299.48} & & failed & 1.29/0.09 & 1.92/0.07 & \textbf{0.95}/\textbf{0.03} \\
& 15 & & 23.44/7125.78 & 3.31/1018.76 & 8.00/4583.68 & \textbf{1.16}/\textbf{444.15} & & failed & 1.66/0.12 & 2.32/0.09 & \textbf{1.08}/\textbf{0.04} \\
& 20 & & 27.21/9281.92 & 4.08/1338.19 & 8.68/6019.29 & \textbf{1.23}/\textbf{581.83} & & failed & 2.01/0.13 & 2.64/0.09 & \textbf{1.23}/\textbf{0.04} \\
         \hline

        \hline
    \end{tabular}
    \label{tab:6}
\end{table*}


\begin{figure}[ht]
    \centering
    \subfloat[EUROC: MH\_04. VIO orientation errors.\label{fig:6a}]{\includegraphics[width=\linewidth]{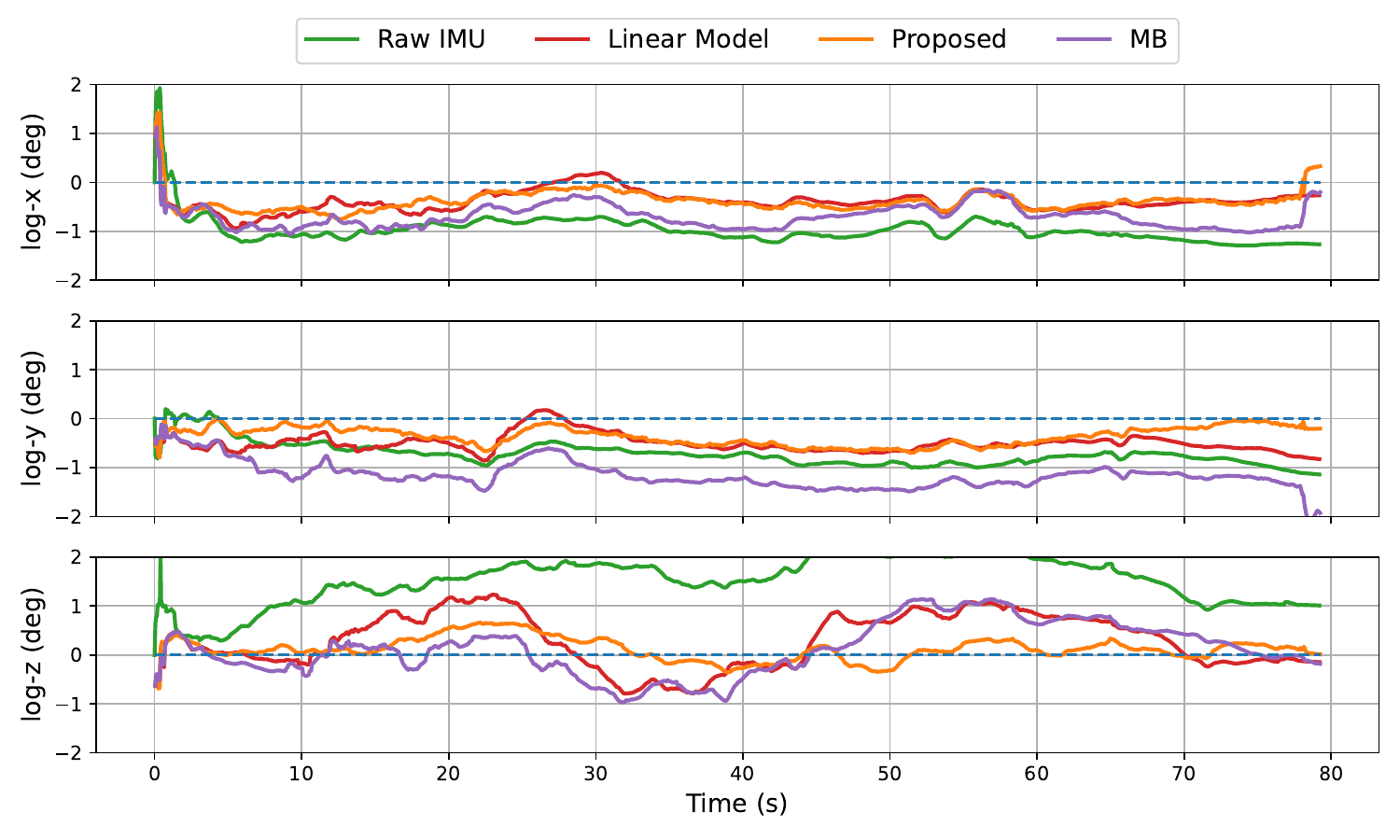}}
    \hfill
    \subfloat[TUM: Room4. VIO orientation errors.\label{fig:6b}]{\includegraphics[width=\linewidth]{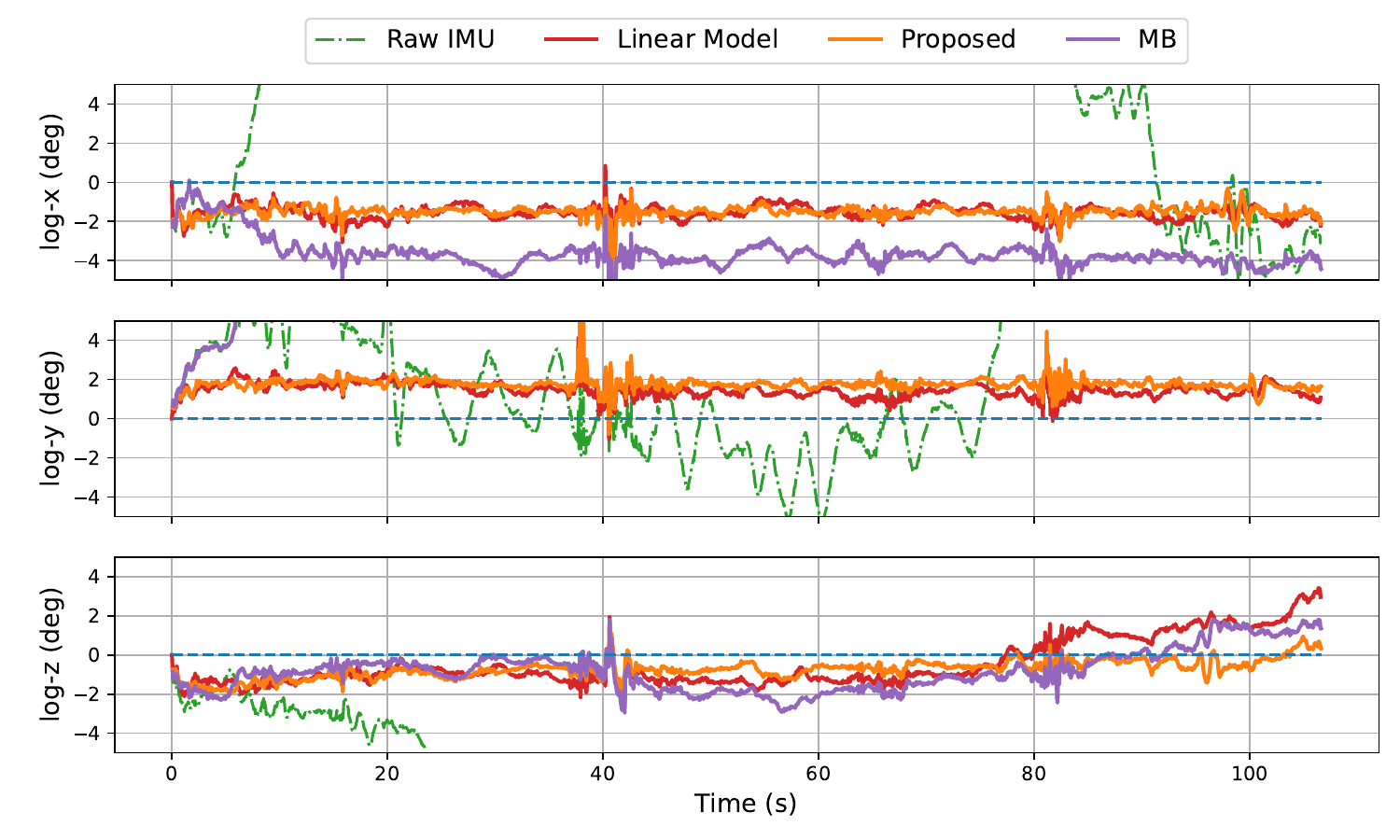}}
    \caption{The coordinates of the \textbf{errors}, $\Log(R_{gt}^\mathsf{T}\hat R)$. The estimates are obtained by VIO with debiased IMU data. The unit is converted to a degree. Closer to zero yields better results.}
    \label{fig:vio_SO3}
    \vspace{-10px}
\end{figure}

\begin{figure}[ht]
    \centering
    \subfloat[EUROC: MH\_04. VIO position errors. \label{fig:7a}]{\includegraphics[width=\linewidth]{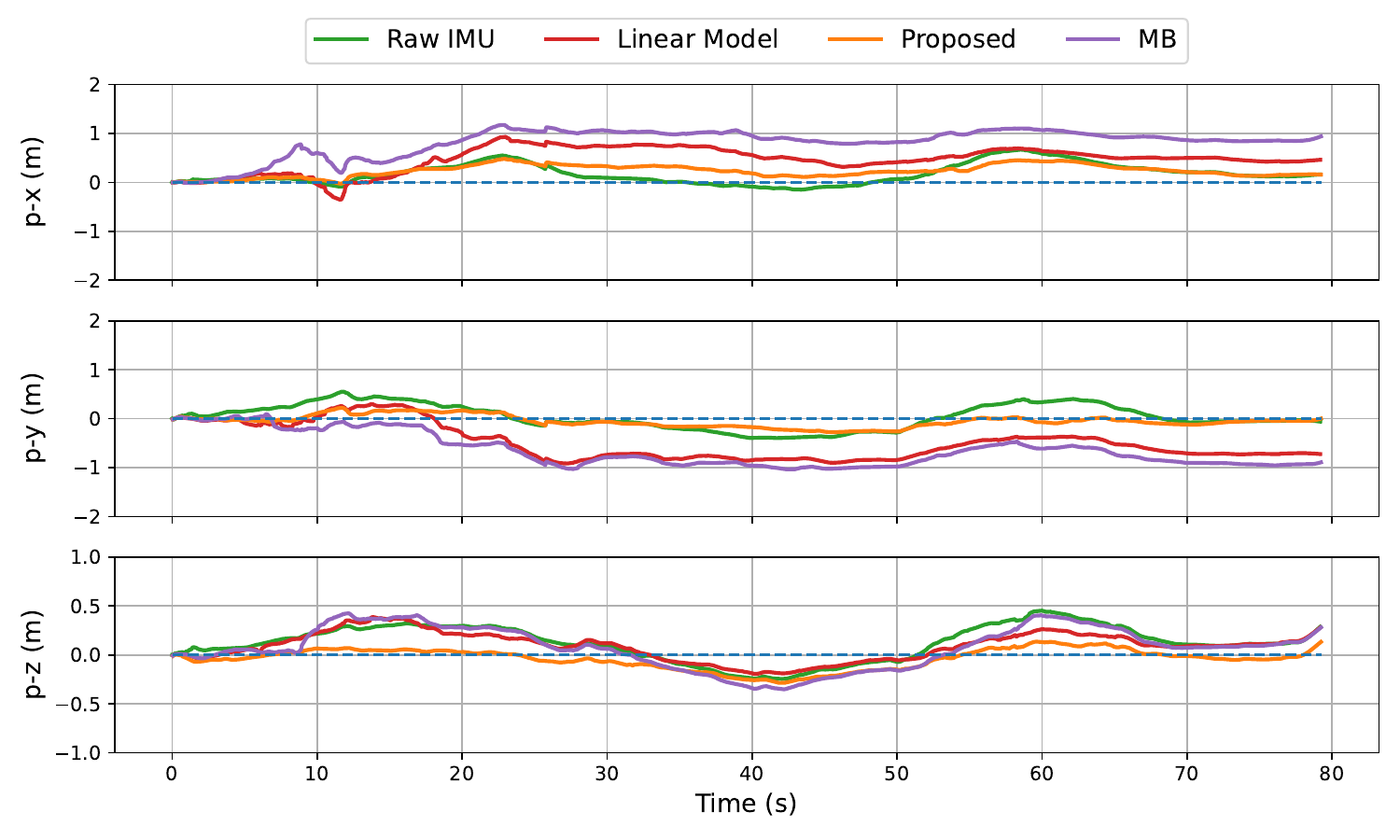}}
    \hfill
    \subfloat[TUM: Room4. VIO position errors. \label{fig:7b}]{\includegraphics[width=\linewidth]{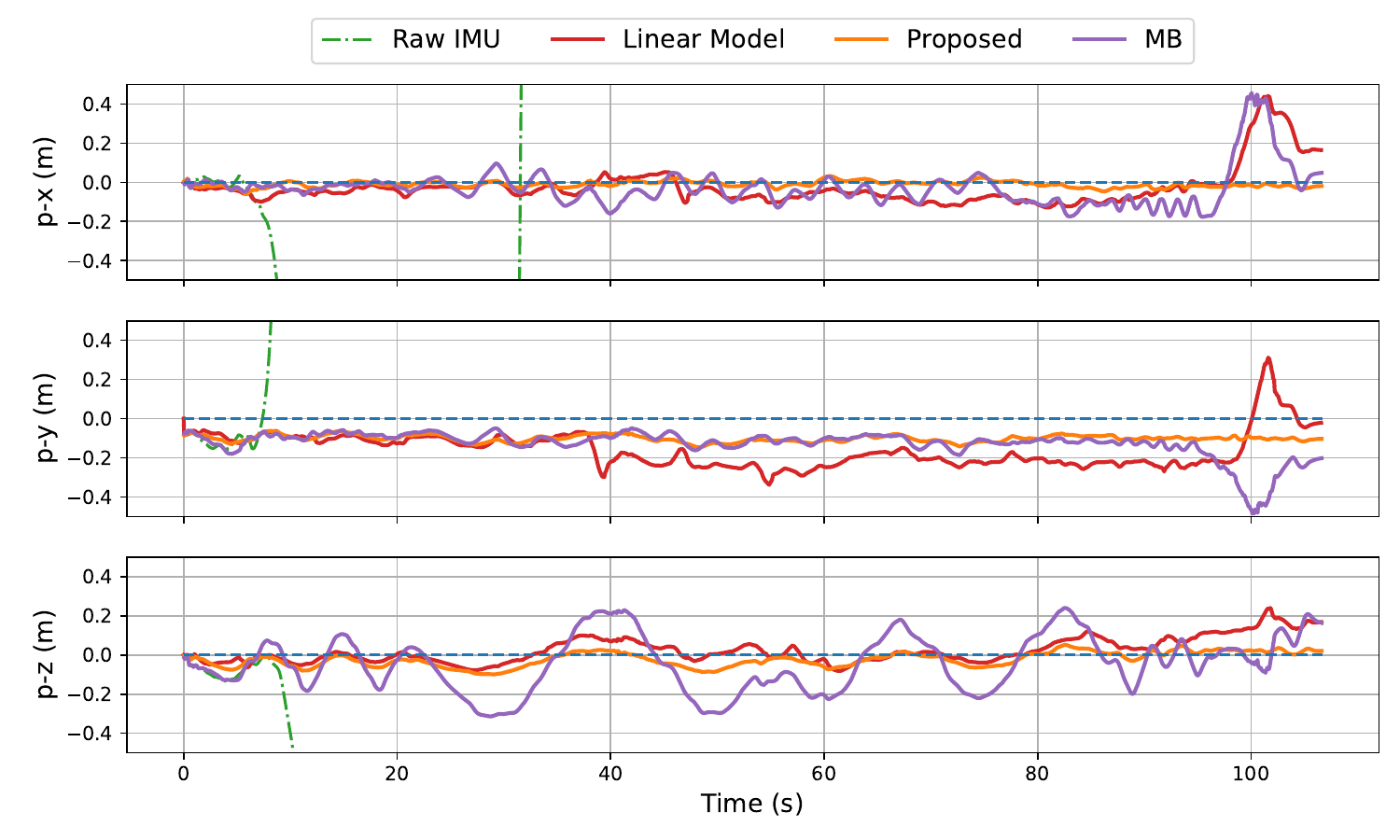}}
    \caption{The position error, $\hat p-p_{gt}$. The estimates are obtained by VIO with debiased IMU data.
    Closer to the ground truth yields better results.}
    \label{fig:vio_position}
\end{figure}

We also employ our network to provide clean IMU measurements as input to OpenVINS \cite{geneva2020openvins}, a VIO algorithm, instead of using raw IMU data. This approach has greater practical relevance. The absolute errors and relative errors are presented in Table \ref{tab:5} and Tab. \ref{tab:6}. Fig.~\ref{fig:vio_SO3} and \ref{fig:vio_position} further plot the orientation errors and positions of different methods in VIO settings. 
The performance improvements with our method are not as significant compared to pure IMU integration since the extra vision sensor can also provide pose estimation itself. It is worth noting that using raw IMU data from the TUM-VI dataset causes the VIO algorithm to fail under our experimental settings, which gives significant errors. 
Although the raw IMU data performance is bad in pure IMU integration, it performs well in the Euroc dataset, primarily due to the bias estimation capabilities of the VIO algorithm. However, this bias estimation is not always reliable, as demonstrated by its failure with raw IMU data in the TUM-VI dataset. From the tables, the proposed method yields better results in both absolute and relative errors within the VIO framework, where the IMU is coupled with other vision sensors. This demonstrates that the accurate IMU measurements provided by our debiasing method can also contribute to improved performance in inertial-based odometry.

\begin{table}[ht]
    \renewcommand{\arraystretch}{1.1}
    \caption{IMU Grades}
    \centering
    \begin{tabular}{ccc}
    \hline

    \hline
    \textbf{Dataset} & \textbf{IMU Model} & \textbf{Grade} \\
    \hline
    EUROC & ADIS16448 & Industrial/tactical \\
    TUM-VI & Bosch BMI160 & Consumer/low-cost \\
    FETCH & Not declared & Consumer/low-cost \\
    \hline

    \hline
    \end{tabular}
    \label{tab:IMUgrade}
\end{table}

\subsection{Debiasing Results on Real-world Experiment}
\begin{figure}
    \centering
    \includegraphics[width=1\linewidth]{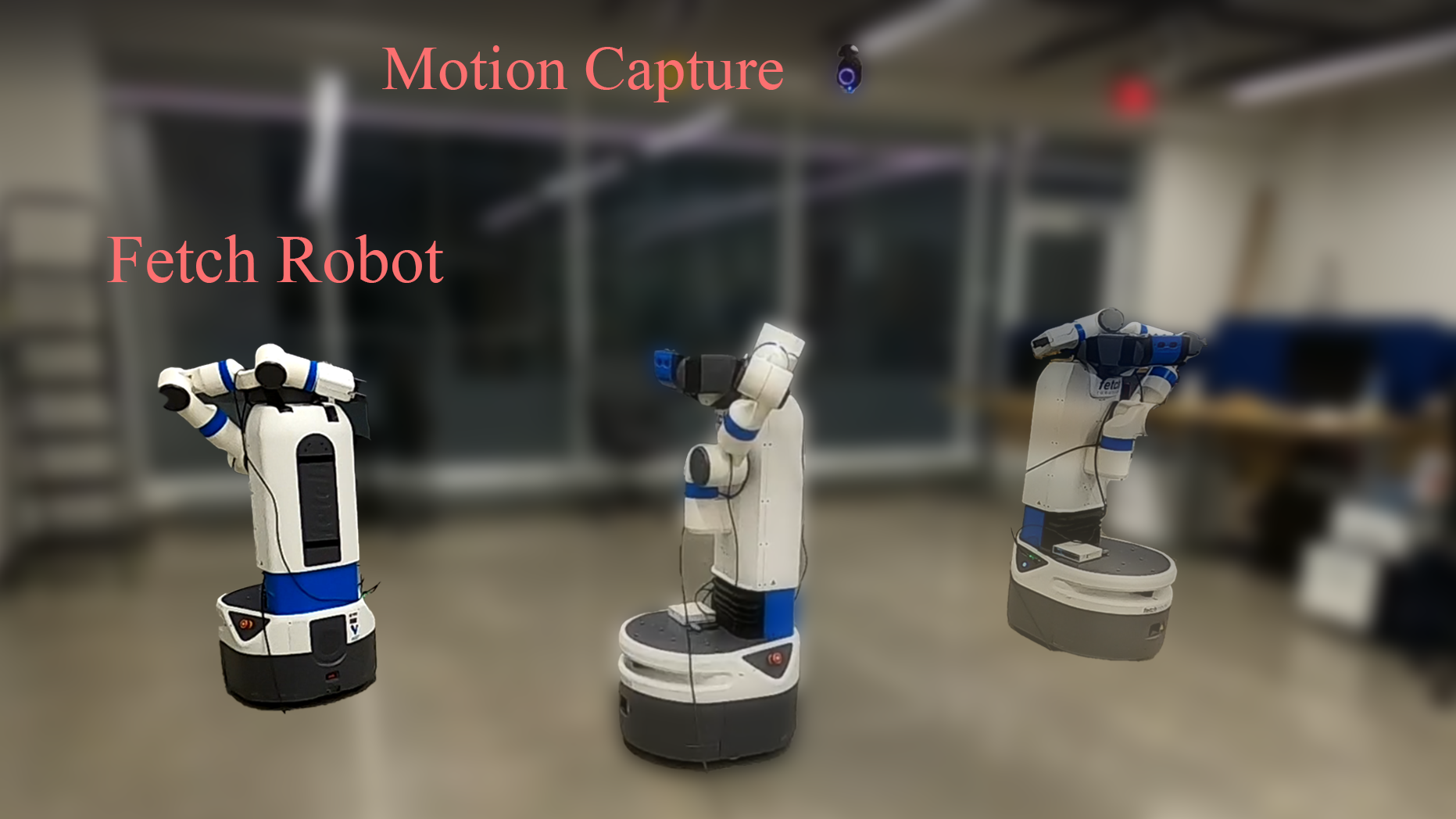}
    \caption{Fetch experiment platform. Fetch is a mobile manipulation, the base equips a low-cost IMU. A motion capture collects the pose ground truth.}
    \label{fig:fetch_fig}
    \vspace{-10px}
\end{figure}

To further demonstrate the generalization capability of the proposed method, we also apply it to a real-world dataset dataset, the FETCH dataset. This dataset is collected from an indoor mobile robot, including six sequences with simple motion patterns, each lasting 30-40 seconds, and one longer trajectory with random motion lasting approximately 90 seconds. We use the longest trajectory for testing and the remaining sequences for training. The experiment platform is shown in Fig. \ref{fig:fetch_fig}.

To demonstrate the practicality of our method, we used the same network parameters on FETCH as for the EUROC dataset. Results in Table \ref{tab:7} show that our method also works well for real-world dataset, consistent with its performance on public datasets. The orientation and velocity estimation is shown in Fig. \ref{fig:1c} and  \ref{fig:2c}, respectively. 
It is surprisingly found that the linear model performs comparably to the proposed method and even achieves better positional accuracy. 
This can be attributed to the simplicity of motion patterns in the FETCH dataset, as we argue that learning-based methods would inevitably capture and leverage such patterns. To support this interpretation, we analyze the performance of the linear model on the more diverse EUROC and TUM-VI datasets, which contain a wider range of motion patterns. In these cases, the linear model performs worse than the learning-based method in both cases, with a particularly poor performance on the TUM-VI dataset, which involves human motion patterns, further validating our hypothesis.
Based on these findings, we recommend first attempting a linear model for IMU calibration to assess whether it meets the desired accuracy requirements. If not, a learning-based approach can then be implemented to achieve better performance.

\begin{table}[ht]
    \renewcommand{\arraystretch}{1.1}
    \caption{Absolute Orientation Error (AOE) and Absolute Position Error (APE) of pure IMU integration for FETCH. Less is better.}
    \centering
    \begin{tabular}{ccccc}
    \hline

    \hline
    \multirow{2}{*}{\textbf{Seq.}} & \multicolumn{4}{c}{\textbf{AOE / APE of IMU Integration} (deg / m)} \\
         &\textbf{Raw IMU} & \textbf{Linear} & \textbf{M.B. \cite{brossard2020denoising}} &\textbf{Proposed} \\
    \hline
        05\_random & 73.28/1139.07 & 2.42/\textbf{119.70} & 2.78/146.45 & \textbf{1.96}/132.57 \\
    \hline

    \hline
    \end{tabular}
    \label{tab:7}
\end{table}

\subsection{Discussion on Implementation}
\textbf{Integration Method:} There is no significant performance difference between the Euler method and more advanced integration methods. This is likely because the high frequency of IMU data and the coupling of integration errors with unknown noise in IMU measurements diminish the impact of the integration method used.

\textbf{Integration Interval Length:} While longer integration windows generally produce smoother bias estimates, simply increasing the window length does not improve accuracy, see Tab. \ref{tab:ablation}.

\textbf{Data Pre-processing:} Pre-processing plays a critical role. We observed that the ground truth in the TUM-VI dataset is noisier than in the EUROC dataset, resulting in noisier initial bias conditions. Applying an appropriate smoother to refine the initial bias conditions leads to better training outcomes.

\textbf{Method Limitations:} Although the proposed method is a device-specific calibration approach with some robustness across varying motion patterns, it still learns motion pattern characteristics during training. As a result, when the available motion pattern information is limited, the method may perform similarly to a simple linear model, losing the advantages in accuracy. Another limitation is that the training time and memory consumption will increase as the integration interval lengthens, which is a common issue in RNN-like networks. The memory consumption issue can be mitigated by developing the adjoint method in the future.

\begin{table*}[ht]
    \renewcommand{\arraystretch}{1.1}
    \centering
    \caption{Ablation Study on Integration Length. We present results on the EUROC dataset showing how performance varies with different integration lengths. AOE and APE are computed from pure integration. Since position drifts quickly, we report APE on velocity instead.}
    \label{tab:ablation}
    \begin{tabular}{ccccccccc}
    \hline

    \hline
     Integration Length N & 4 (0.2~s) & 8 (0.4~s)&  12 (0.6~s) & 16 (0.8~s) & 32 (0.16~s) & 48 (0.24~s) & 64 (0.32~s) & 96 (0.48~s)\\
     \hline
     AOE (deg)& 91.59 & 12.53 & 2.50 & 2.71 & 2.62 & 2.45 & 3.48 & 2.70 \\
APE (velocity) (m/s) & 257.89 & 32.74 & 19.94 & 17.75 & 20.82 & 18.02 & 16.36 & 18.58 \\
Training Time (s)& 135.62 & 300.83 & 460.36 & 573.63 & 1223.78 & 1823.72 & 2483.10 & 3610.63 \\
GPU Memory (MB)& 590.73 & 418.34 & 476.53 & 533.20 & 762.34 & 988.30 & 1217.71 & 1671.22 \\
     \hline

    \hline
         
    \end{tabular}
    \includegraphics[width=.9\linewidth]{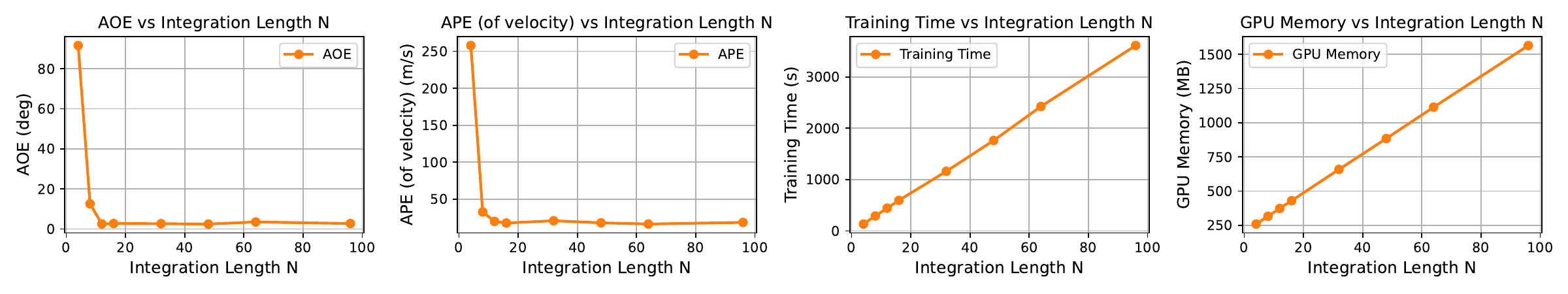}
\end{table*}

Additionally, we provide an ablation study on integration length, with results shown in Tab. \ref{tab:ablation} on the EUROC dataset. The accuracy is poor with short integration lengths mainly due to noises. The accuracy can improve as the integration length increases, however, the accuracy saturates around $N = 10$. Since both training time and GPU memory consumption grow roughly linearly with the integration length, we set $N = 16$ in our experiments to balance accuracy and resource usage. Our findings are consistent with \cite{chen2018neuralode}, where GPU memory consumption scales as $\mathcal{O}(N)$ when the adjoint method is not used.

\section{Conclusion}
\label{sec:conclusion}
In this work, we propose a learning-based method to obtain accurate angular velocity and acceleration using only raw IMU measurements. The proposed method explicitly learns the bias dynamics of the IMU using a NODE framework without the need for the bias ground truth. With the known bias dynamics, we can compensate for the bias and obtain a clean IMU measurement. 
Experiments on two popular datasets and one real-world experiment demonstrate that the proposed method outperforms the existing approach and has practical meaning.

In the future, this work can be extended to incorporate covariance propagation through the learned dynamics, allowing integration with any existing IMU-based odometry systems.
Although the current work does not fully address bias propagation with covariance, one promising application is to use the unbiased IMU data within the \emph{invariant extended Kalman filter} framework, which effectively eliminates the nonlinearities in IMU kinematics \cite{barrau2017Invariant,barrau2018invariantKalmanfiltering,paul2018invariantsmoothing}.
Additionally, the backpropagation of the NODE on Lie algebra formulation can be extended into more efficient adjoint method.

\section*{Acknowledgment}
Ben Liu would like to thank Hongbo Zhu for his invaluable assistance in this work.
Wei Zhang was supported by the Guangdong Science and Technology Program under Grant No. 2024B1212010002 and 2019QN01X793 for this work.

\appendix
\subsection{Proof of Theorem \ref{thm:1}}
We need to show $\dot R_t$ meets the differential equation and $R_t(0)=R_0$. First, we need the derivative of the exponential map for $\mathrm{SO}(3)$ \cite[Theorem 5.4]{hall2015liegroup_complete}: 
\begin{equation}
    \frac{\D \exp(\xi^\times)}{\D t} = \exp(\xi^\times)\left(\sum_{k=0}^{\infty}(-1)^k\frac{\ad_{\xi^\times}^k}{(k+1)!}\cdot( \frac{\D \xi^\times}{\D t})\right)
\end{equation}
where $\ad_\xi\in\R^{3\times 3}$ is the \emph{adjoint map}. This can be expressed in further simplified formulation \cite[eq. 7.77a]{barfoot2017state}:
\begin{equation}
    \frac{\D \Exp(\xi)}{\D t} = \Exp(\xi)(J_r(\xi)\dot\xi)^\times
\end{equation}
which can be substituted into $R_t=R_0\Exp(\xi_t)$, then
\begin{equation}
    \dot R_t = R_0\Exp(\xi_t)(J_r(\xi) J_r^{-1}(\xi_t)\omega_t)^\times = R_t\omega^\times
\end{equation}
with the initial condition: $R_t(0)=R_0\Exp(\bm{0})=R_0$. \hfill $\square$


{
\small\balance
\bibliographystyle{plainnat}
\bibliography{ref}
}

\end{document}